\documentclass[authoryear,preprint,review,3p,times]{elsarticle}

\usepackage{amssymb}
\usepackage{latexsym}
\usepackage[nointegrals]{wasysym}
\usepackage{url}
\usepackage{xcolor}
\usepackage{amsmath}
\usepackage{booktabs}
\usepackage{balance}
\usepackage{adjustbox}
\usepackage{threeparttable}
\usepackage{subfigure}
\usepackage{algorithm}
\usepackage{algorithmic}
\usepackage{framed,multirow}

\definecolor{airforceblue}{rgb}{0.36, 0.54, 0.66}
\definecolor{ballblue}{rgb}{0.004, 0.50, 0.69}
\definecolor{newcolor}{rgb}{.8,.349,.1}

\usepackage[colorlinks]{hyperref}
\AtBeginDocument{%
  \hypersetup{
    citecolor=ballblue,
    linkcolor=ballblue,   
    urlcolor=ballblue}}

\newcommand{\pl}[0]{}

\journal{Computers in Biology and Medicine}

\begin{document}

\begin{frontmatter}



\title{Long-tailed Medical Diagnosis with Relation-aware Representation Learning and Iterative Classifier Calibration}


\author[1]{Li Pan\fnref{fn1}}
\author[2]{Yupei Zhang\fnref{fn1}}
\fntext[fn1]{Equal contribution.}
\author[3]{Qiushi Yang}
\author[4]{Tan Li}
\author[5]{Zhen Chen\corref{cor1}}
\cortext[cor1]{Corresponding author.}\ead{zchen.francis@gmail.com}

\affiliation[1]{{Department of Pathology, The University of Hong Kong}}
\affiliation[2]{{Department of Clinical Neurosciences, University of Cambridge}}
\affiliation[3]{{Department of Electrical Engineering, City University of Hong Kong}}
\affiliation[4]{{Department of Computer Science, The Hang Seng University of Hong Kong}}
\affiliation[5]{{CAIR, HKISI, CAS}}

\begin{abstract}
Recently computer-aided diagnosis has demonstrated promising performance, effectively alleviating the workload of clinicians. However, the inherent sample imbalance among different diseases leads algorithms biased to the majority categories, leading to poor performance for rare categories. Existing works formulated this challenge as a long-tailed problem and attempted to tackle it by decoupling the feature representation and classification.  Yet, due to the imbalanced distribution and limited samples from tail classes, these works are prone to biased representation learning and insufficient classifier calibration. To tackle these problems, we propose a new Long-tailed Medical Diagnosis (LMD) framework for balanced medical image classification on long-tailed datasets. In the initial stage, we develop a Relation-aware Representation Learning (RRL) scheme to boost the representation ability by encouraging the encoder to capture intrinsic semantic features through different data augmentations. In the subsequent stage, we propose an Iterative Classifier Calibration (ICC) scheme to calibrate the classifier iteratively. This is achieved by generating a large number of balanced virtual features and fine-tuning the encoder using an Expectation-Maximization manner. The proposed ICC compensates for minority categories to facilitate unbiased classifier optimization while maintaining the diagnostic knowledge in majority classes. Comprehensive experiments on three public long-tailed medical datasets demonstrate that our LMD framework significantly surpasses state-of-the-art approaches. The source code can be accessed at \href{https://github.com/peterlipan/LMD}{https://github.com/peterlipan/LMD}.
\end{abstract}


%%\item We present a novel method to tackle long-tailed problems in medical image diagnosis
%%\item The proposed RRL module efficiently enhances the representation learning of encoders
%%\item The ICC iteratively calibrates the classifier with abundant balanced virtual features
%%\item Experiments on three medical datasets illustrates the effectiveness of our method

\end{frontmatter}


\section{Introduction}
\label{sec:introduction}
Over recent years, computer-aided diagnosis has achieved remarkable success, presenting the ability to reduce the burden on clinicians \citep{srinidhi2021deep,zhang2024unified,yang2022semi,chen2021super}. 
However, common diseases have a disproportionately higher number of samples than the rare ones in real-world medical datasets, due to the inherent class imbalance caused by different target diseases \citep{yang2020rethinking,chen2022personalized}.
This class imbalance has been recognized as a long-tailed issue, where a small number of head classes have abundant samples, whereas tail classes consist of limited instances \citep{esteva2017dermatologist}. This long-tailed distribution misleads model training towards majority categories, severely degrading the diagnosis performance \citep{cui2019class}.

To combat the long-tails, numerous current approaches have generally sought to modify the data distribution by reducing the samples of the dominant classes \citep{buda2018systematic}, increasing the samples of the minority classes \citep{zhang2014rwo}, or reweighting the contribution of various classes in the optimization \citep{cui2019class}. Nevertheless, these resampling-based methods suffer from performance decreases on certain long-tailed datasets because the entire information capacity of the dataset is either the same or even decreases \citep{zhang2021bag, yang2020rethinking}. Two-stage approaches have driven recent progress in the long-tailed classification. Initially, these approaches train the model on the whole imbalanced dataset, followed by a second stage where the classifier is fine-tuned using rebalancing strategies to address class imbalance \citep{kang2019decoupling, cao2019learning}. By decoupling encoders and classifiers' training, two-stage approaches can leverage all training data to improve representation learning of encoders and calibrate biased classifiers on the rebalanced subset. 

\begin{figure}[t!]
\centering
\makebox[\linewidth][c]{\includegraphics[width=.75\linewidth]{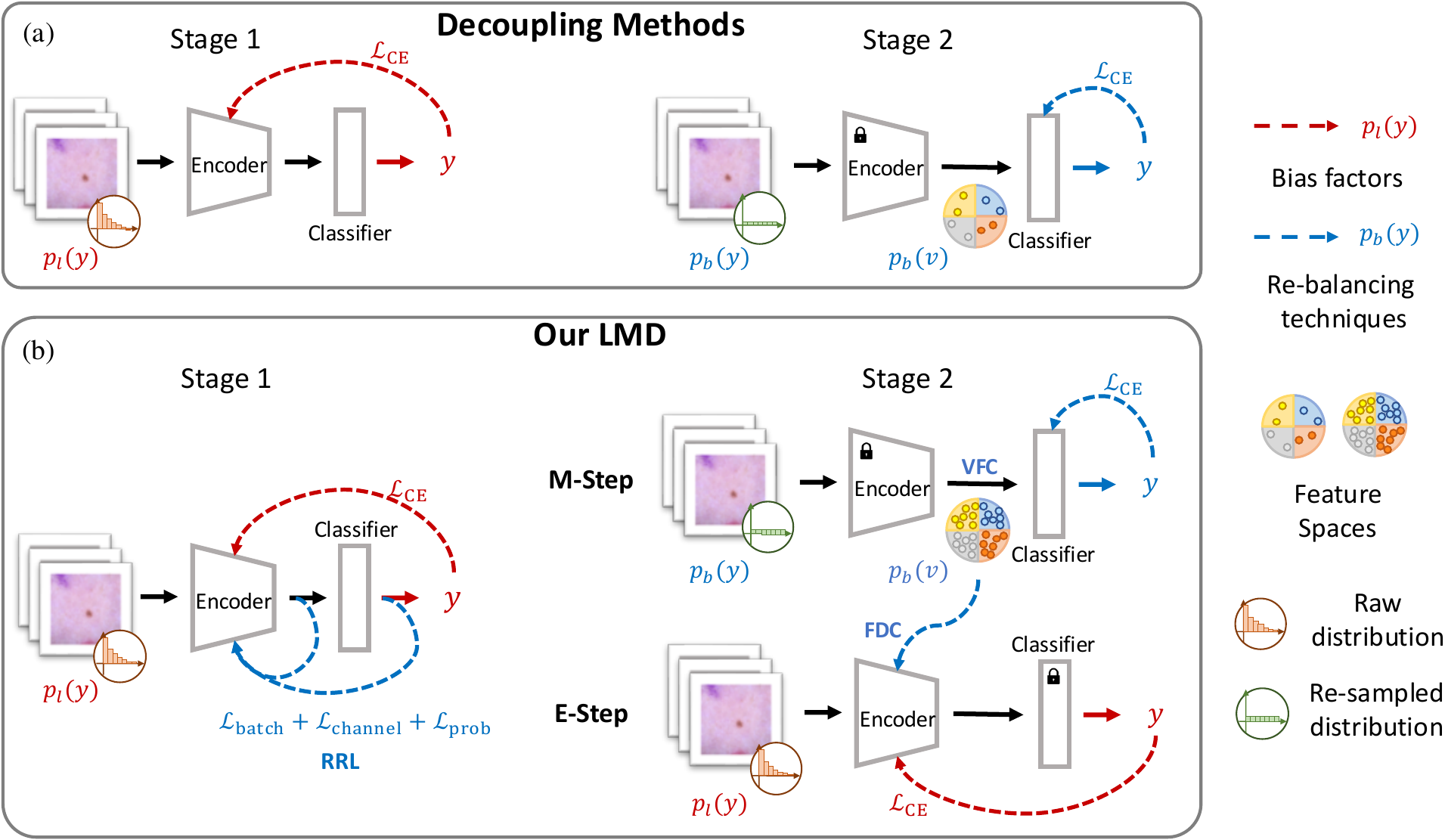}}
\caption{Comparison of (a) the decoupling methods \citep{kang2019decoupling} and (b) our LMD in long-tailed medical image diagnosis. \pl{The LMD enhances the representation learning of encoders and promotes classifier calibration with virtual features and iterative training.}} \label{fig:preliminary}
\end{figure}

Despite the success of decoupling methods, the imbalanced classification performance on tail classes is still worth improving \citep{zhang2021deep, li2022flat}. We identify two primary challenges among previous decoupling methods as follows.
First, as illustrated in Fig. \ref{fig:preliminary}, decoupling approaches train the model on the long-tailed dataset in the first stage, which is inadequate and imbalanced for the representation learning due to the scarcity of tail samples \citep{liu2020deep}. \cite{marrakchi2021fighting} attempted to tackle this issue by boosting encoders' representation capacity in the first stage via contrastive learning. However, the effectiveness of this approach relies on the definition of meaningful positive and negative pairs, which in turn necessitates a substantial amount of samples. To enhance and balance the representation learning, as shown in Fig. \ref{fig:preliminary}, we propose a new representation learning scheme in the first stage, encouraging the model to learn intrinsic semantic information from input images with distinct augmentations.

The second challenge of the decoupling approaches arises in the second stage, where the pre-trained encoder is frozen and the biased classifier is fine-tuned \citep{kang2019decoupling, li2022flat}. Traditional rebalancing methods, like reweighting and resampling, are devised to retrain an unbiased classifier. However, these ad-hoc rebalancing methods balance the classifications at the expense of losing the knowledge from the head classes, e.g., resampling-based approaches tend to overlook the information contained in the head classes \citep{li2024feature}, and reweighting-based approaches cannot effectively address the class imbalance using simple coefficient adjustments \citep{wang2021seesaw}.
Hence, developing a new decoupling approach that can efficiently calibrate the classifier with abundant and balanced data in the second stage is crucial.
Moreover, the decoupled training strategy of the two stages may lead the classifier calibration to sub-optimum.
As discussed above, \cite{kang2019decoupling} disentangled the training process of the encoder and classifier to alleviate the bias within the classifier. However, since the classifier is trained on the feature space constructed by the encoder, the two components remain intertwined, leading to suboptimal results \citep{ur2017adaptive, eryilmaz2020optimization}. To tackle this challenge, \cite{li2022flat} introduced a regularization term in the first stage to encourage the first-stage optimization to converge at a more stable optimum, preventing the second-stage training from escaping the local optima of the first stage. Nonetheless, the optima achieved in the first stage does not necessarily satisfy the optimization target of the second stage, and any updates made to the second-stage model can affect the optimization target of the first stage \citep{zhou2023class, wang2022balanced}. 
To address the above issues, as shown in Fig. \ref{fig:preliminary}, we propose an Expectation-Maximization classifier calibration strategy, which iteratively fine-tunes the encoder and calibrates the classifier with abundant virtual features.

In this work, we introduce the LMD framework to balance the recognition performance on long-tailed medical datasets. Specifically, to enhance the model's representation learning from limited tail class samples, we design a Relation-aware Representation Learning approach for encouraging the semantic feature extraction from images with different data perturbations. In stage two, to improve the classification performance, we devise an Iterative Classifier Calibration (ICC) strategy, which iteratively fine-tunes the encoder and classifier in an Expectation-Maximization manner. During the Maximization step, we present the Virtual Features Compensation (VFC) to compensate for tail classes by generating virtual features under the multivariate Gaussian distribution. During the Expectation step, we propose the Feature Distribution Consistency (FDC) loss to fine-tune the encoder without being affected by the biased data distribution.
By these means, the proposed LMD framework can calibrate biases that exist in the encoder and classifier, and construct a balanced, representative latent space to enhance classification performances, especially on rare diseases. The conducted experiments demonstrate that our LMD framework is superior to state-of-the-art approaches on public medical imaging datasets. Our contributions are four-fold:
\begin{itemize}
\item To tackle the long-tailed problem in medical image diagnosis, we propose a novel Long-tailed Medical Diagnosis framework, by addressing the imbalanced representation learning and insufficient classifier calibration in decoupling learning.
\item To enhance the representation ability of encoders, especially on the tail classes, we propose Relation-aware Representation Learning, which constrains the consistency of encoders regarding different data perturbations from multiple views.
\item We propose the Iterative Classifier Calibration, which calibrates the classifier with balanced virtual features and Feature Distribution Consistency using an Expectation-Maximization approach.
\item We conduct experiments on three long-tailed datasets, including \emph{ISIC-2019-LT}, \emph{ISIC-Archive-LT}, and \emph{Hyper-Kvasir}, which prove the superiority of our LMD framework in medical image diagnosis with long-tails.
\end{itemize}

A preliminary version of this work has been published in MICCAI 2023 \citep{pan2023combat}. In this work, we have made a significant extension with the following highlights: 1) We propose the Iterative Classifier Calibration (ICC) to fine-tune the encoder and calibrate the classifier, along with the Feature Distribution Consistency (FDC) loss to address the imbalance; 2) Compared with the conference work \citep{pan2023combat}, we implement the balanced feature distribution estimation to combat the imbalance in the Virtual Features Compensation (VFC) and illustrate its efficiency with extensive experiments; 3) Besides experiments on dermoscopy images, we conduct extensive experiments to enhance the comprehensive validation, including experiments on gastrointestinal (GI) dataset, comparison with existing long-tail works and detailed ablation studies.

\section{Related Work}
% 自然场景long tail
\subsection{Long-tailed Classification}
Deep neural networks have demonstrated promising performance on various computer vision benchmarks, encompassing image classification  \citep{luo2024surgplan,chen2023star,chen2023surgical,yang2023hierarchical} and image segmentation \citep{yang2022d,zhu2023feddm,chen2024sam}.
However, real-world datasets usually follow a long-tailed class distribution, where most labels are associated with only a few samples but others are associated with only a few samples \citep{li2022long}. 
Such imbalanced distribution makes the data-sensitive deep learning models trained by naive likelihood maximization strategy biased towards the majority classes, leading to poor model performance on the minority classes \citep{lu2023label}. This impaired performance on the tail classes has hindered the implementation of deep learning models in real-world scenarios, becoming an increasing concern \citep{jin2023long}.

To tackle the challenge of class imbalance, a straightforward way is to resample the original dataset to retain a class-balanced subset, including over-sampling the tail classes \citep{more2016survey}, under-sampling the head classes \citep{buda2018systematic}, or sampling each class with the uniform probability \citep{kang2019decoupling}. Some studies \citep{lin2017focal, wang2021seesaw} propose to reweight the contribution of different classes to the loss function gradient to reach a balanced solution. \cite{lin2017focal} assigned a higher weight to misclassified examples that are hard to classify, while down-weighting easy examples that are correctly classified, to improve the performance on tail classes. \cite{cao2019learning} adjusted the margin between the decision boundary and training samples based on the label distribution, moving the boundary towards rare classes. \cite{wang2021seesaw} adaptively rebalanced positive and negative gradients for each category to mitigate the punishments to tail classes as well as compensate for the risk of misclassification caused by diminished penalties.
Yet, these re-weighting techniques improve the performance of tail classes at the cost of degradation on the head classes.

\subsection{Long-tails in Medical Imaging}
\pl{With rapid advancements, deep learning methods have demonstrated a strong capability in medical image classification tasks \citep{almalik2022self, liu2021medical, chen2020joint,chen2022instance}, highlighting the ability of computer-aided diagnosis and helping to alleviate the workload of clinicians \citep{zhao2022reasoning,chen2021diagnose,pan2024focus}}.
Meanwhile, the medical datasets are naturally imbalanced due to the scarcity of disease samples, causing the same long-tailed problems \citep{yang2020rethinking}. In the medical field, where constructing datasets is costly and diagnostic accuracy is crucial, addressing the challenges posed by long-tailed data is of utmost importance \citep{islam2021review}.

% existing methods except the decoupling
To mitigate the long-tailed problem in medical imaging, \cite{khushi2021comparative} explored a set of resampling-based methods, including under-sampling majority categories and over-sampling minority categories, to construct balanced subsets from the original dataset. 
\pl{\cite{chen2023pcct} proposed a novel class-balanced triplet sampler to alleviate the class imbalance in representation learning.}
\cite{rezaei2020addressing} proposed weighted cross-entropy loss, which manually adjusts the weight of the components of cross-entropy loss to address the long-tailed problem in medical image classification.
\cite{galdran2021balanced} performed instance-based and class-based re-sampling of the training data and mixed up the two sets of samples to construct a more balanced dataset.
\cite{ju2022flexible} incorporated a curriculum learning module with resampling methods to query new samples with per-class difficulty-aware sampling probability. However, these resampling approaches tend to undersample the head classes and lack the mechanism to synthesize new data for the tail classes, thereby limiting the model performance on the majority classes while providing marginal improvement for the minority groups.

\subsection{Decoupling Learning for Long-tails}
Despite the long-tailed problem causing performance degradation, \cite{tang2020long} pointed out that representation learning of encoders can still benefit from imbalanced data.
\cite{yang2020rethinking} proposed that even the imbalanced labeled data can be leveraged to boost the model's representation ability, but also emphasized that this may reduce classification performance due to classifier bias.  
To retain the visual representation ability of the encoder and alleviate the bias in the classifier, \cite{kang2019decoupling} disentangled the training process of the encoder and the classifier, which first trains the encoder on the whole dataset and then fine-tunes the classifier on frozen features under class-balanced sampling.

With the success of decoupling methods in the computer vision field \citep{zhou2023class, nam2023decoupled,chen2023medical}, recent long-tailed medical image classification tasks have adopted this two-stage training strategy.
In particular, \cite{chen2021self} conducted unsupervised learning in the first stage to eliminate the impact of label space and fine-tune the model on the class-balanced dataset to address the long-tailed problem.
\cite{marrakchi2021fighting} employed supervised contrastive learning in the first stage, which separates the feature space into different clusters by minimizing the distance between samples from the same class and maximizing the distance between samples from different classes, to boost the representation learning of the encoder.
\cite{li2022flat} proposed a flat-aware optimization strategy to approach a flatter optimum in the first stage, which better coordinates the training of the two stages. Nevertheless, these decoupling methods still suffer from imbalanced representation learning in the first stage and insufficient classifier calibration in the second stage, which can lead to suboptimal results. Different from existing decoupling methods, our LMD framework enhances representation learning with the multi-view relation-aware consistency strategy and iteratively calibrates the classifier with abundant virtual features.

\begin{figure*}[t]
\centering
\makebox[\textwidth][c]{\includegraphics[width=.99\textwidth]{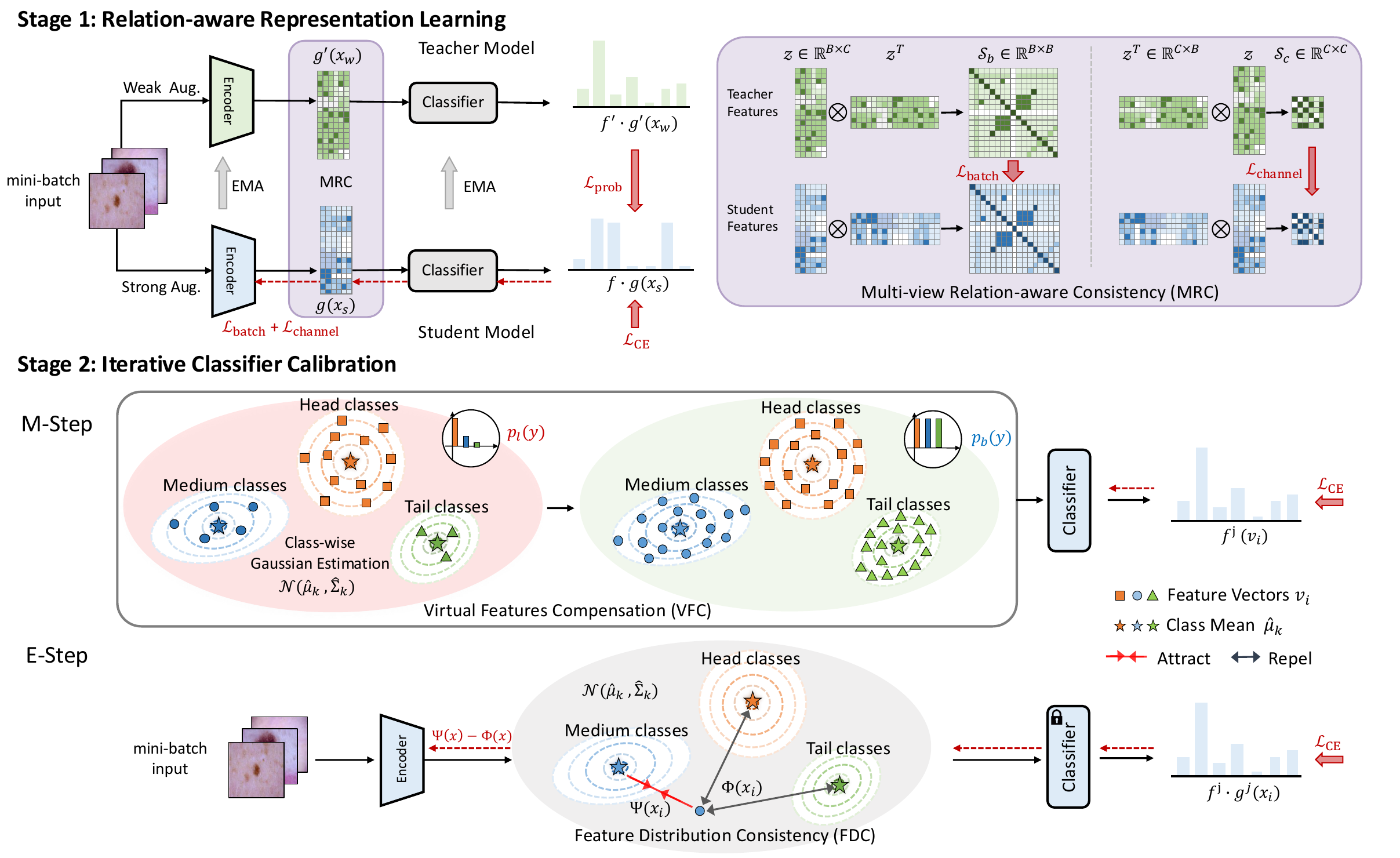}}
\caption{The illustration of our LMD framework. (a) In the Relation-aware Representation Learning, we enhance encoder's the representation learning ability with the MRC module on imbalanced datasets. (b) In the Iterative Classifier Calibration, we calibrate the classifier with abundant virtual features generated by VFC during the Maximization step and fine-tune the encoder with FDC during the Expectation step.} \label{fig:framework}
\end{figure*}

\section{Methodology}

\subsection{Preliminaries}
We start by revisiting the training strategy of the decoupling \citep{kang2019decoupling} in long-tailed image recognition.
As shown in Fig. \ref{fig:preliminary}, to combat the long-tailed distribution $p_l(y)$, decoupling disentangles the training process of the encoder $g$ and the classifier $f$. \pl{In the first stage, \cite{kang2019decoupling} jointly trained the parameters of classifier $\theta_f$ and encoder $\theta_f$ on the imbalanced dataset as follows}:
\begin{equation} \label{eq:stage1}
\begin{split}
    \theta_g^*, \hat{\theta_f} \ &= \underset{\theta_g, \theta_f}{\arg\min} \ -\sum_{i=1}^{N}\log P(y_i \mid g(\mathbf{x}_i; \theta_g), \theta_f) \\
    &= \ \underset{\theta_g, \theta_f}{\arg\min} \ -\sum_{i=1}^{N} \log\frac{P_l(y_i\mid \theta_f) \ P_l(g(\mathbf{x}_i; \theta_g) \mid y, \theta_f)}{P_l(g(\mathbf{x}_i; \theta_g)\mid \theta_f)},
\end{split}
\end{equation}
where $N$ denotes the number of samples in the dataset, $\mathbf{x}$ represents the input images, $y$ indicates the labels.
\pl{The whole training process is conducted on the ill distribution $P_l(y)$, leading to biased representation learning on the tail classes.}

In the second stage, \cite{kang2019decoupling} sampled each class $k$ with an equal probability $p_k=1/K$, and $K$ means the number of classes in the dataset, to construct a class-balanced subset $P_b(x)$. Then, the classifier is retrained on the unbiased dataset using cross-entropy loss as follows:
\begin{equation} \label{eq:stage2}
\begin{split}
    \theta_f^* \ = \ \underset{\theta_f}{\arg\min}& \ -\sum_{i=1}^{M}\log P(y_i \mid \mathbf{v}_i, \hat{\theta_f}), \\
    &{\rm w.r.t.} \ \mathbf{v}_i \ = \ g(\mathbf{x}_i, \theta_g^*),
\end{split}
\end{equation}
where $\theta_g^*$ is the parameters of the encoder $g$ gained in the first stage, $M$ represents the number of samples after resampling, and $\mathbf{v}_i$ means the feature vector of sample $\mathbf{x}_i$. Note that resampling does not generate new instances, i.e., $M \leq N$, and the number of resampled data is constrained by the tail classes, which leads to a lack of training samples and ultimately decreases classification performance. Furthermore, as shown in Eq. \eqref {eq:stage1} and Eq. \eqref {eq:stage2}, the optimization of $ \theta _f $ and $ \theta _g $ is coupled. The update of the classifier simultaneously changes the optimization target of $ \theta _g^* $, which in turn changes the feature space $ \{ \mathbf{v}_i \} _ { i=1 } ^M $. This change leads to a sub-optimal performance as it affects the optimization target of $ \theta _f $.

\subsection{Overview}
As depicted in Fig.~\ref{fig:framework}, our LMD framework follows the decoupling strategy \citep{kang2019decoupling,zhou2020bbn} to tackle long-tailed challenges. In stage one, we introduce Relation-aware Representation Learning to enhance the encoder $g$'s representation capability through the Multi-view Relation-aware Consistency (MRC) module. In stage two, we devise the Iterative Classifier Calibration strategy to calibrate the classifier $f$ and fine-tune the encoder $g$ using an Expectation-Maximization approach. During the Maximization step, we calibrate the classifier $f$ by generating a large number of balanced virtual features with VFC. During the Expectation step, we fine-tune the encoder $g$ under the Feature Distribution Consistency loss. By enhancing the representation learning with RRL and calibrating the classifier with ICC, our LMD framework can achieve balanced and effective training on long-tailed medical datasets.

\subsection{Relation-aware Representation Learning}
As discussed above, the encoder's representation learning is insufficient, especially on the tail classes \citep{zhang2021bag, zhang2021deep}. To enhance representation learning, we devise Relation-aware Representation Learning, which aims to help the encoder capture the semantic characteristics of input images through various data augmentations. In detail, we propose a student model $f \cdot g$ with strong augmented images as inputs $\mathbf{x}_s$ and replicate a teacher neural network $f{'} \cdot g{'}$ with weak augmented images as inputs $\mathbf{x}_w$. The MRC module constrains student and teacher models to ensure consistency across different perturbations of the same input. The teacher model's parameters are updated using an exponential moving average \citep{tarvainen2017mean} of the student model's parameters.

To encourage the student model to learn from the imaging patterns of inputs while decreasing the impact of imbalanced label distribution, we propose a novel multi-view constraint to promote consistency between the two models. For the same input image under different augmentation processes, We encourage the teacher and student to achieve identical predictions:
\begin{gather}\label{eq:prob}
    \mathcal{L}_{\rm prob} \ = \ \frac{1}{B}  {\rm KL}(f\cdot g(\mathbf{x}_{s}) \ , \ f{'}\cdot g{'}(\mathbf{x}_{w})),
\end{gather}
where $\rm KL(\cdot,\cdot)$ represents the Kullback–Leibler divergence that quantifies the difference between two input distributions. To further facilitate consistent representations of the identical image with minor perturbations, we propose MRC that directly guides encoder training by maximizing the sample-wise and channel-wise similarities between the encoders of the teacher and student. Given the Gram matrix as $\mathcal{S}$, we first define the relationship among samples and among channels as $\mathcal{S}_{b}(\mathbf{z}) \ = \  \mathbf{z} \cdot \mathbf{z}^\intercal$ and $\mathcal{S}_{c}(\mathbf{z}) \ = \  \mathbf{z}^\intercal \cdot \mathbf{z}$, respectively.
The vector $\mathbf{z}=g(\mathbf{x}_{s}) \in \mathbb{R}^{B \times C}$ indicates the output feature map of the last layer of the encoder $g(\cdot)$. $B$ and $C$ are the number of samples and channels, respectively. $\mathcal{S}_b(\mathbf{z})$ represents the relationships among samples, and $\mathcal{S}_c(\mathbf{z})$ measures the similarities among channels. We further calculate the sample-wise and channel-wise consistency as follows:
\begin{align}
    \mathcal{L}_{\rm sample}\ &= \ \frac{1}{B} (\mathcal{S}_{b}(g(\mathbf{x}_{s})) \ - \ \mathcal{S}_{b}(g{'}(\mathbf{x}_{w})))^2, \\
    \mathcal{L}_{\rm channel}\ &= \ \frac{1}{C}  (\mathcal{S}_{c}(g(\mathbf{x}_{s})) \ - \ \mathcal{S}_{c}(g{'}(\mathbf{x}_{w})))^2. \label{eq:channel}
\end{align}
Additionally, to ensure accurate classification of images and avoid potential collapse of the optimization process, the cross-entropy loss $\mathcal{L}_{\rm CE} \ = \ \frac{1}{B} L(f\cdot g(\mathbf{x}_{w}),\ y)$, where $y$ represents the label, is also adopted. \pl{We summarize the overall optimization target as $\mathcal{L}_{\rm stage1} \ = \  \mathcal{L}_{\rm CE} + \lambda_1 (\mathcal{L}_{\rm sample} +  \mathcal{L}_{\rm channel} + \frac{1}{2} \mathcal{L}_{\rm prob})$, where $\lambda_1$ is the hyperparameter that balances the trade-off among each loss term and will be discussed in the ablation study.}
% $\mathcal{L}_{\rm CE}(\cdot)$ represents the cross-entropy loss function. 
\pl{The proposed RRL module enhances the representation capabilities of encoders by promoting consistent representations $g(\mathbf{x)}$ for images with various augmentations $\{\mathbf{x}_s, \mathbf{x}_w\}$ from multiple views $\{\mathcal{L}_{\rm sample}, \mathcal{L}_{\rm channel}, \mathcal{L}_{\rm prob}\}$. RRL thus alleviates the class imbalances in representation learning, facilitating balanced feature distributions in the latent space, and ultimately benefiting balanced classification.}

% Virtual Features Compensation for classifier calibration (VFC)
\subsection{Iterative Classifier Calibration}
The decoupling methods \citep{kang2019decoupling, zhang2019study} froze the encoder to maintain the feature representations and fine-tuned the classifier on the balanced dataset constructed by resampling technologies to mitigate the bias within the classifiers. However, optimizing the two components separately may lead to a suboptimum \citep{ur2017adaptive, eryilmaz2020optimization}. To address this problem, we design an Iterative Classifier Calibration scheme, which iteratively fine-tunes the encoder and calibrates the classifier using an Expectation-Maximization strategy to approach the global optimum. During the Expectation step, we fine-tune the encoder with the FDC loss. During the Maximization step, we calibrate the classifier with virtual features generated by VFC.

\subsubsection{Virtual Features Compensation}
Decoupling methods \citep{kang2019decoupling} disentangle the training process of encoders and classifiers to alleviate the imbalance within classifiers while preserving the representation capabilities of encoders. Nevertheless, as shown in Fig. \ref{fig:preliminary}, to eliminate the bias within the classifier, existing decoupling techniques resample the imbalanced dataset by discarding samples from the head classes, leading to insufficient learning. To address this issue, we propose the VFC module to generate balanced virtual features $v_k \in \mathbb{R}^{R \times C}$ for each category $k$ under multivariate Gaussian distributions. Unlike existing resampling methods, the virtual features maintain inter-class correlations and intra-class semantic information while enabling balanced feature distribution. For the $k$-th class, we first estimate the class-specific multivariate Gaussian distribution $\mathcal{N}(\boldsymbol{\mu}_k, \boldsymbol{\Sigma}_k)$ was:
\begin{equation}\label{eq:gaussian}
\begin{gathered}
    \boldsymbol{\mu}_k \ = \ \frac{1}{N_k}\sum_{\mathbf{x} \in X_k} g(\mathbf{x}),  \\
    \boldsymbol{\Sigma}_k \ = \ \frac{1}{N_k - 1} \sum_{\mathbf{x} \in X_k} (g(\mathbf{x}) - \boldsymbol{\mu}_k)^\intercal (g(\mathbf{x}) - \boldsymbol{\mu}_k),
\end{gathered}
\end{equation}
where $X_k$ represents the group of samples belonging to category $k$, $g(\cdot)$ is initialized as the encoder trained in the first stage, and $N_k$ denotes the number of samples in class $k$. For each class, We randomly sample $R$ feature vectors under the class-specific multivariate Gaussian distribution to construct a balanced latent space, as $\{V_{k}\in \mathbb{R}^{R \times C}\}_{k=1}^{K}$. After obtaining the virtual features for each class, we use them to augment the original training set. Specifically, we replace the original feature vectors with the sampled virtual features to form a balanced feature space. The impact of the number of sampled features for each class $R$ will be discussed in the ablation study.

\subsubsection{Maximization Step}
In the proposed Maximization step, the encoder is frozen and the classifier is trained to maximize the classification performance in the feature space. Specifically, we first estimate the multivariate Gaussian distribution of the features generated by the encoder. To eliminate the bias inside the distribution estimation caused by the imbalanced label space, we adopt a class-balanced sampling strategy \citep{zhang2021bag} as $p_k \ = \ 1 / K, \  \mathbb{E}[\hat{N}_k] \ = \ N / K$, where $p_k$ represents the probability of class $k$ to be selected, $\hat{N}_k$ indicates the number of instances from class k after resampling. With each class to have a uniform probability of being selected, the expectation of mean and covariance for each class can be estimated as:
\begin{equation}\label{eq:m_gaus}
\begin{gathered}
    \mathbb{E}[\hat{\boldsymbol{\mu}}_k] \ = \ \frac{K}{N}\sum_{\mathbf{x} \in \hat{X}_k} \mathbb{E}[g(\mathbf{x})], \\
    \mathbb{E}[\hat{\boldsymbol{\Sigma}}_k] \ = \ \frac{K}{N - K} \sum_{\mathbf{x} \in \hat{X}_k} \mathbb{E}[(g(\mathbf{x}) - \boldsymbol{\mu}_k)^\intercal (g(\mathbf{x}) - \boldsymbol{\mu}_k)],
\end{gathered}
\end{equation}
where $\hat{X}_k$ refers to the subset of class $k$ after resampling, $\hat{\boldsymbol{\mu}}_k$ and $\hat{\boldsymbol{\Sigma}}_k$ indicates the estimated mean and covariance of the multivariate Gaussian distribution of the $k$-th class. As illustrated in Eq. \eqref{eq:m_gaus}, the estimated mean and covariance are irrelevant to the number of samples for class $k$, i.e., $N_k$, indicating a more balanced statistics of the feature distribution. At each iteration of the Expectation-Maximization optimization, the estimated mean and covariance are updated using the exponential moving average \citep{tarvainen2017mean}. With the unbiased mean and covariance of each class, we generate an equal number of virtual features for each class using VFC to construct a balanced feature space and train the classifier as follows:
\begin{equation}\label{eq:m}
\begin{aligned}
    \theta_f^{j} \ = \ \underset{\theta_f}{\arg\min} \ -\sum_{i=1}^{RK}\log P(y_i \mid \mathbf{v}_i, \theta_f), \\
    \mathcal{L}^{M}_{\rm stage2} \ = \ \frac{1}{RK} \sum_{k=1}^{K} \sum_{\mathbf{v}_i\in V_{k}} \mathcal{L}_{\rm CE}(f^j(\mathbf{v}_i), y),
\end{aligned}
\end{equation}
where $K$ is the number of classes in the raw dataset, $f^j(\cdot)$ indicates the classifier at the $j$-th iteration, and $f^0(\cdot)$ is re-initialized to mitigate the bias within the first-stage classifier. Different from the existing decoupling methods which have limited training samples due to under-sampling as illustrated in Eq. \eqref{eq:stage2}, the proposed VFC can generate abundant virtual features for each class with an equal number, boosting training of the classifier.

\begin{figure}[t!]
\begin{algorithm}[H] 
    \caption{The pipeline of LMD} \label{pseudocode}
    \begin{algorithmic}[1]
        \REQUIRE{Images $X = \{x_i\}_{i=1}^{N}$; Labels $Y = \{y_i\}_{i=1}^{N}$; Encoder $g(\cdot)$; Classifier $f(\cdot)$}
        \ENSURE{Predictions $\hat{Y} = \{\hat{y}_i\}_{i=1}^{N}$}
        
        \textbf{Stage 1: Relation-aware Representation Learning}
        \STATE $X_s \gets$ strong-augment($X$); $X_w \gets$ weak-augment($X$)
        \STATE $f^{'}(\cdot) \gets f(\cdot)$; $g^{'}(\cdot) \gets g(\cdot)$
        \WHILE{$\mathcal{L}_{\rm stage1}$ does not converge}
        \STATE Calculate $\mathcal{L}_{\rm sample}$, $\mathcal{L}_{\rm channel}$, $\mathcal{L}_{\rm prob}$ using Eq. \eqref{eq:prob} to \eqref{eq:channel}
        \STATE \pl{$\mathcal{L}_{\rm stage1} \ = \  \mathcal{L}_{\rm CE} + \lambda_1 (\mathcal{L}_{\rm sample} + \mathcal{L}_{\rm channel} + \frac{1}{2} \mathcal{L}_{\rm prob})$}
        \STATE Update student model $f(\cdot)$ and $g(\cdot)$
        \STATE $f^{'}(\cdot) \stackrel{\text{\tiny EMA}}{\gets} f(\cdot)$; $g^{'}(\cdot) \stackrel{\text{\tiny EMA}}{\gets} g(\cdot)$
        \ENDWHILE

        \textbf{Stage 2: Iterative Classifier Calibration}
        \STATE Initialize $f^0(\cdot)$; $g^0(\cdot) \gets g(\cdot)$; $j \gets 0$
        \WHILE{$j < J$}
        \STATE Freeze encoder $g^j(\cdot)$ and unfreeze classifier $f^j(\cdot)$
        \STATE Estimate class-wise $\hat{\mu}_k$ and $\hat{\Sigma}_k$ using Eq. \eqref{eq:gaussian}
        \STATE Randomly sample $R$ samples from each class $k$ under distribution $N(\hat{\mu}_k, \hat{\Sigma}_k)$ 
        \STATE Calculate the M step loss $\mathcal{L}_{\rm stage2}^{\rm M}$ using Eq. \eqref{eq:m} and update $f^{j}(\cdot)$ 

        \STATE Freeze classifier $f^j(\cdot)$ and unfreeze encoder $g^j(\cdot)$
        \STATE Calculate the E step loss $\mathcal{L}_{\rm stage2}^{\rm E}$ using Eq. \eqref{eq:e} and update $g^j(\cdot)$ 
        \STATE $j \gets j + 1$
        \ENDWHILE

        \STATE Prediction $\hat{y}_i \ = \ f^j \cdot g^j (x_i)$
    \end{algorithmic}
\end{algorithm}
\end{figure}

\subsubsection{Expectation Step} 
To preserve the knowledge inside the classifier, in the expectation step, we freeze the classifier $f(\cdot)$ and train the encoder $g(\cdot)$ to calculate the expected distribution of the features as follows:
\begin{equation}
    \theta_g^{j} \ = \ \underset{\theta_g}{\arg\min} -\sum_{i=1}^{N}\log P(g(\mathbf{x}_i, \theta_g) \mid \mathbf{x}_i, \theta_f^{j}).
\end{equation}
As discussed in Eq. \eqref{eq:stage1}, the imbalanced data brings bias to the training of the encoder.
To avoid the encoder being re-contaminated by the imbalanced label distributions and to make use of all training samples for improved representation learning, we propose a new regularizer based on the multivariate Gaussian distribution. Intuitively, given the unbiased estimation of the mean and covariance as shown in Eq. \eqref{eq:m_gaus}, we encourage the model to learn the feature representations where features are close to their class means and far away from mean vectors of other classes. We formulate the attraction $\Psi(\mathbf{x})$ and repulsion $\Phi(\mathbf{x})$ as follows:
\begin{equation}\label{eq:attraction}
    \Psi(\mathbf{x}) = \frac{1}{B}\sum_{i=1}^{B} (g^j(\mathbf{x}_i) - \hat{\boldsymbol{\mu}}_{k_i}) \hat{\Sigma}_{k_i} (g^j(\mathbf{x}_i) - \hat{\boldsymbol{\mu}}_{k_i})^\intercal,
\end{equation}
\begin{equation}\label{eq:repulsion}
    \Phi(\mathbf{x}) = \frac{1}{B}\sum_{i=1}^{B} \frac{1}{K-1}\sum_{k\neq k_i} (g^j(\mathbf{x}_i) - \hat{\boldsymbol{\mu}}_{k}) \hat{\Sigma}_{k} (g^j(\mathbf{x}_i) - \hat{\boldsymbol{\mu}}_{k})^\intercal,
\end{equation}
where $\mathbf{x}_i$ denotes the $i$-th data sample, $k_i$ indicates the class of the $i$-th sample, $g^j(\cdot)$ represents the encoder at $j$-th iteration, and $g^0(\cdot)$ is initialized as the encoder trained in the first stage. In particular, $\Psi(\mathbf{x})$ quantifies the average Mahalanobis distances \citep{de2000mahalanobis} between the samples and corresponding class means, and $\Phi(\mathbf{x})$ measures the average distances between the samples and other class mean vectors. By minimizing $\Psi(\mathbf{x})$ and maximizing $\Phi(\mathbf{x})$, the feature vectors are pulled towards their class means and pushed away from the mean vectors of other classes. To avoid potential collapsing solutions, we implement a cross-entropy loss and formulate the Feature Distribution Consistency (FDC) loss as follows:
\begin{equation}\label{eq:e}
    \mathcal{L}_{\rm stage2}^{\rm E} \ = \ \lambda_e(\Psi(\mathbf{x}) - \Phi(\mathbf{x})) + \frac{1}{B}\sum_{i=1}^{B}\mathcal{L}_{\rm CE}(f^{j}\cdot g^j(\mathbf{x}_i), y_i),
\end{equation}
where $\lambda_e$ indicates the trade-off between the regularizer and cross-entropy loss, which will be discussed in the ablation study.  By regularizing the cross-entropy loss with the proposed constraint in the Expectation step, the encoder can be optimized regarding the updates of the classifier in the previous iteration without involving the imbalance bias. 
\pl{With an abundant set of balanced virtual features $\{V_{k}\in \mathbb{R}^{R \times C}\}_{k=1}^{K}$ generated by the VFC module, the proposed ICC iteratively calibrates the classifier using the FDC constraint in an Expectation-Maximization framework, thereby alleviating imbalances and promoting balanced classification performance across all classes.}

\subsection{Algorithm Pipeline}
The pipeline of our proposed LMD framework is summarized in Algorithm \ref{pseudocode}, which includes the Relation-aware Representation Learning and the Iterative Classifier Calibration.
We first randomly initialize the student model $f\cdot g$ and the teacher model $f^{'}\cdot g^{'}$ as the same. In the first stage, we train the two models with strong $X_s$ and weak $X_w$ augmentations, respectively, according to the loss functions $\mathcal{L}_{\rm prob}$, $ \mathcal{L}_{\rm sample}$, $\mathcal{L}_{\rm channel}$, and $\mathcal{L}_{\rm CE}$ defined in Eq. \eqref{eq:prob} to \eqref{eq:channel}. In the second stage, we design an Expectation-Maximization optimization schedule. In the $j$-th iteration of the expectation step, we estimate the expected distribution of the features regarding the classifier $f^j(\cdot)$ with the loss function $\mathcal{L}_{\rm stage2}^{\rm E}$ defined in Eq. \eqref{eq:e}. In the $j$-th iteration of the Maximization step, we fine-tune the biased classifier in the balanced latent space generated by the encoder $g^j(\cdot)$ with the loss function $\mathcal{L}_{\rm stage2}^{\rm M}$ defined in Eq. \eqref{eq:m}.
% The algorithm pipeline is shown in \ref{pseudocode}.

\begin{table*}[t!]
\centering
\caption{Comparisons on the \emph{Hyper-Kvasir} dataset.}
\label{tab:kvasir}
\resizebox{.95\textwidth}{!}{
\begin{tabular}{l|c|c|c|c|c}
\toprule
\multicolumn{6}{c}{\emph{Hyper-Kvasir}}                              \\
\hline
\hline
Methods & AUC (\%) & F1 (\%) & Kappa (\%) & Precision (\%) & Recall (\%) \\
\hline
    CE  & 95.61  & 46.62 & 78.54 & 47.78 & 47.53 \\
    RS \citep{zhang2021bag}  & 95.53 & 60.11  & 85.70 & 59.69 & 60.81 \\
    Focal loss \citep{lin2017focal} & 98.46  & 58.91 & 85.99 & 60.33 & 58.35 \\
    LDAM-RS \citep{cao2019learning} & 94.77  & 49.96 & 81.08 & 50.08 & 50.57 \\
    CB-Focal \citep{cui2019class} & 95.37  & 58.27 & 85.25 & 58.24 & 58.48 \\
    Decoupling \citep{kang2019decoupling} & 94.68  & 60.53 & 86.07 & 60.92 & 60.36 \\
    Seesaw \citep{wang2021seesaw} & 98.09  & 59.49 & 84.99 & 60.59 & 59.13 \\
    CICL \citep{marrakchi2021fighting} & 97.41  & 57.59 & \underline{91.69} & 58.99 & 57.47 \\
    Bal-Mixup \citep{galdran2021balanced} & 96.83  & 58.59 & 91.13 & 60.37 & 58.10 \\
    CB+WD+Max \citep{alshammari2022long} & 97.41  & 59.62 & 91.42 & 60.32 & 59.59 \\
    FCD \citep{li2022flat} & 96.12  & 57.28 & 85.13 & 57.45 & 61.91 \\
    FS \citep{ju2022flexible} & 95.80  & 57.34 & 86.41 & 58.09 & 58.39 \\
    CC-SAM \citep{zhou2023class} & 98.84  & \underline{61.72} & 87.19 & \underline{61.60} & 63.40 \\
    GCL \citep{li2024adjusting} & 98.62  & 60.59 & 86.26 & 61.34 & 62.85 \\

    \hline
    % Ours $w/o$ EM    &  68.15 & 63.29  &  67.13 & 66.19\\
    LMD $w/o$ RRL & 95.99  & 59.63 & 89.93 & 58.55 & 61.19 \\
    LMD $w/o$ ICC & 98.53  & 60.94 & 88.95 & 59.88 & 63.62 \\
    LMD $w/o$ VFC & \underline{98.92}  & 57.52 & 91.03 & 59.48 & 57.33 \\
    \pl{LMD $w/o$ FDC} &  \pl{98.68}  & \pl{61.51}  & \pl{87.98} & \pl{60.42}  & \pl{\underline{63.89}}  \\
    \textbf{LMD} & \textbf{98.96}  & \textbf{62.99} & \textbf{92.43} & \textbf{62.35} & \textbf{67.27} \\
\toprule
\end{tabular}}
\end{table*}

\section{Experiments}
\subsection{Datasets}
\noindent \textbf{ISIC Datasets.} To verify the performance on the long-tailed medical image classification tasks, we construct two imbalanced datasets from the ISIC \citep{tschandl2018ham10000} following \citep{ju2022flexible}. Specifically, we construct the \emph{ISIC-2019-LT} dataset, including $8$ diagnostic classes of dermoscopic images, as the long-tailed version of the ISIC 2019 challenge \citep{codella2018skin, combalia2019bcn20000}. We generate the subset from the Pareto distribution \citep{cui2019class} using the formula $N_{c}=N_{0}(r^{-(k-1)})^c$, where the imbalance factor $r=N_{0}/N_{k-1}$ is defined by the ratio of the sample volume of the head class $N_{0}$ to that of the tail class $N_{k-1}$. For the \emph{ISIC-2019-LT}, we used three different imbalance factors: $r= \{100, 300, 500\}$. Additionally, the \emph{ISIC-Archive-LT} dataset \citep{ju2022flexible} is constructed from the ISIC Archive with a larger imbalance factor of approximately $r=1000$ and includes dermoscopic images across $14$ categories. These two datasets are randomly split into training, validation, and testing sets in a 7:1:2 ratio.

\begin{table*}[t]
\centering
\caption{Comparisons on the \emph{ISIC-Archive-LT} dataset.}
\label{tab:archive}
\resizebox{.95\textwidth}{!}{
\begin{tabular}{l|c|c|c|c}
\toprule
\multicolumn{5}{c}{\emph{ISIC-Archive-LT}} \\
\hline
\hline
\multirow{2}{*}{Methods} & \multicolumn{4}{c}{BACC (\%)} \\
\cline{2-5}
 & Head & Medium & Tail & Overall \\
\hline
    CE 	& 71.26 $\pm$ 0.19 & 50.39 $\pm$ 0.12 & 26.72 $\pm$ 0.33 & 46.21 $\pm$ 0.30 \\
    RS \citep{zhang2021bag}  & 71.75 $\pm$ 0.20 & 49.30 $\pm$ 0.15 & 33.63 $\pm$ 0.31 & 49.00 $\pm$ 0.28 \\
    Focal loss \citep{lin2017focal} & 70.96 $\pm$ 0.22 & 41.59 $\pm$ 0.09 & 29.43 $\pm$ 0.36 & 44.77 $\pm$ 0.31 \\
    LDAM-RS \citep{cao2019learning} & 71.34 $\pm$ 0.23 & 40.88 $\pm$ 0.17 & 30.21 $\pm$ 0.30 & 43.90 $\pm$ 0.29 \\
    CB-Focal \citep{cui2019class} & 52.03 $\pm$ 0.18 & 54.95 $\pm$ 0.15 & 36.66 $\pm$ 0.35 & 46.27 $\pm$ 0.26 \\
    Decoupling \citep{kang2019decoupling} & 72.15 $\pm$ 0.19 & 48.39 $\pm$ 0.13 & 31.18 $\pm$ 0.33 & 47.90 $\pm$ 0.29 \\  
    Seesaw \citep{wang2021seesaw} & 71.60 $\pm$ 0.20 & 49.37 $\pm$ 0.08 & 42.97 $\pm$ 0.35 & 52.98 $\pm$ 0.27 \\  
    CICL \citep{marrakchi2021fighting} & \textbf{76.33 $\pm$ 0.16} & 51.94 $\pm$ 0.15 & 35.01 $\pm$ 0.30 & 51.65 $\pm$ 0.28 \\
    Bal-Mixup \citep{galdran2021balanced} & 69.91 $\pm$ 0.24 & 38.99 $\pm$ 0.24 & 27.80 $\pm$ 0.29 & 43.03 $\pm$ 0.31 \\
    CB+WD+Max \citep{alshammari2022long} & 64.73 $\pm$ 0.16 & 29.74 $\pm$ 0.25 & 24.87 $\pm$ 0.39 & 37.65 $\pm$ 0.34 \\
    FCD \citep{li2022flat} & 70.59 $\pm$ 0.21 & 51.76 $\pm$ 0.18 & 41.33 $\pm$ 0.31 & 52.87 $\pm$ 0.26 \\
    FS \citep{ju2022flexible} & 68.14 $\pm$ 0.19 & 54.62 $\pm$ 0.21 & 44.81 $\pm$ 0.26 & 51.89 $\pm$ 0.24 \\
    CC-SAM \citep{zhou2023class} & 67.72 $\pm$ 0.13 & \underline{55.03 $\pm$ 0.16} & \underline{52.60 $\pm$ 0.24} & \underline{57.61 $\pm$ 0.19} \\
    GCL \citep{li2024adjusting} & 70.24 $\pm$ 0.20 & 43.45 $\pm$ 0.14 & 42.38 $\pm$ 0.33 & 51.06 $\pm$ 0.22 \\

    \hline
    % Ours $w/o$ EM    &  68.15 & 63.29  &  67.13 & 66.19\\
    LMD $w/o$ RRL & 68.45 $\pm$ 0.19 & 51.34 $\pm$ 0.20 & 42.23 $\pm$ 0.33 & 53.93 $\pm$ 0.29 \\
    LMD $w/o$ ICC & 65.47 $\pm$ 0.18 & 53.26 $\pm$ 0.15 & 44.59 $\pm$ 0.24 & 54.38 $\pm$ 0.23 \\
    LMD $w/o$ VFC & \underline{72.32 $\pm$ 0.21} & 49.24 $\pm$ 0.13 & 38.45 $\pm$ 0.36 & 53.56 $\pm$ 0.32 \\
    \pl{LMD $w/o$ FDC} & \pl{65.89 $\pm$ 0.23} & \pl{53.45 $\pm$ 0.37} & \pl{41.27 $\pm$ 0.22} & \pl{53.47 $\pm$ 0.26} \\

    \textbf{LMD} & 63.79 $\pm$ 0.14 & \textbf{61.05 $\pm$ 0.12} & \textbf{60.59 $\pm$ 0.31} & \textbf{61.63 $\pm$ 0.21} \\
\toprule
\end{tabular}
}
\end{table*}

\noindent \textbf{Hyper-Kvasir Dataset.} Hyper-Kvasir \citep{Borgli2020} is a comprehensive dataset of gastrointestinal (GI) images obtained from endoscopy videos. Endoscopy is the preferred method for examining abnormalities and diseases of the digestive system. This dataset comprises 10,662 images categorized into 23 classes with a long-tailed distribution. The imbalance factor $r$ of this dataset is 171. Following \citep{yue2022toward}, we randomly split the dataset into training, validation, and testing sets as 8:1:1.

\subsection{Implementation Details}
Our LMD framework is implemented using the PyTorch library \citep{paszke2019pytorch}. We use ResNet-18 \citep{he2016deep}, pre-trained on ImageNet \citep{deng2009imagenet}, as the backbone. All experiments are conducted on four NVIDIA GTX 1080 Ti GPUs with a batch size of 128. Images are resized to $224 \times 224$ pixels. In the first stage, we use Stochastic Gradient Descent (SGD) with a learning rate of 0.01 as the optimizer. Strong augmentation \citep{info11020125} is applied using random flip, optical blur, random rotate, color jitter, grid dropout, and normalization strategies. For weak augmentation, only random flip and normalization strategies are used. In the second stage, SGD with a learning rate of $1 \times 10^{-5}$ is employed for classifier optimization, and a learning rate of $1 \times 10^{-6}$ is used for encoder optimization. \pl{The loss weight $\lambda_1$ in the first stage is set to $10$.}

We have compared our LMD framework with state-of-the-art methods on the long-tailed medical image classification task, including (i) baselines: ResNet-18 \citep{he2016deep} pre-trained on the ImageNet \citep{deng2009imagenet} with cross-entropy loss (CE), class-balanced resampling method (RS) \citep{zhang2021bag}; (ii) recent loss reweighting methods: Focal loss \citep{lin2017focal}, Class-Balancing (CB) losses \citep{cui2019class}, LDAM loss \citep{cao2019learning}, and seesaw loss \citep{wang2021seesaw} (iii) recent studies in computer vision: Decoupling \citep{kang2019decoupling}, CB+WD+Max \citep{alshammari2022long}, GCL \citep{li2024adjusting}, and CC-SAM \citep{zhou2023class}. (iv) recent works for long-tailed medical image recognition: Bal-Mixup \citep{galdran2021balanced}, CICL \citep{marrakchi2021fighting}, FCD \citep{li2022flat}, and FS \citep{ju2022flexible}.

\begin{table*}[t]
\centering
\caption{Comparisons with state-of-the-art approaches on the \emph{ISIC-2019-LT} dataset.}
\label{tab:isic19}
\resizebox{.95\textwidth}{!}{
\begin{tabular}{l|c|c|c|c|c|c}
\toprule
\multicolumn{7}{c}{\emph{ISIC-2019-LT}} \\
\hline
\hline
\multirow{2}{*}{Methods} & \multicolumn{2}{c|}{$r=100$} & \multicolumn{2}{c|}{$r=300$} & \multicolumn{2}{c}{$r=500$} \\
\cline{2-7}
 & AUC (\%) & BACC (\%) & AUC (\%) & BACC (\%) & AUC (\%) & BACC (\%) \\
\hline
    CE  & 89.35 & 51.22 & 87.54 & 47.16 & 83.89 & 39.05 \\
    RS \citep{zhang2021bag}  & 94.93 & 59.65 & 91.46 & 55.59 & 86.08 & 42.63 \\
    Focal loss \citep{lin2017focal} & \underline{94.98} & 57.47 & 92.55 & 48.49 & \underline{92.11} & 46.98 \\
    LDAM-RS \citep{cao2019learning} & 93.40 & 56.09 & 90.26 & 48.37 & 88.96 & 48.14 \\
    CB-Focal \citep{cui2019class} & 92.28 & 61.51 & 88.61 & 52.85 & 86.47 & 48.89 \\
    Decoupling \citep{kang2019decoupling} & 94.97 & 60.56 & 92.79 & 53.89 & 90.91 & 52.44 \\
    Seesaw \citep{wang2021seesaw} & 94.85 & 62.37 & 91.93 & 55.09 & 90.09 & 52.26 \\
    CICL \citep{marrakchi2021fighting} & 94.50 & 58.43 & 92.15 & 54.27 & 89.53 & 51.73 \\
    Bal-Mixup \citep{galdran2021balanced} & 93.84 & 51.91 & 93.40 & 49.66 & 90.61 & 40.77 \\
    CB+WD+Max \citep{alshammari2022long} & 93.40 & 58.04 & 91.35 & 50.07 & 85.83 & 48.47 \\
    FCD \citep{li2022flat} & 93.52 & 63.49 & 92.07 & \underline{56.23} & 89.64 & 50.36 \\
    FS \citep{ju2022flexible} & 92.21 & 62.03 & 90.89 & 53.95 & 88.62 & 48.29 \\
    CC-SAM \citep{zhou2023class} & 94.30 & \underline{65.36} & 91.71 & 55.26 & 89.45 & \underline{52.66} \\
    GCL \citep{li2024adjusting} & 93.76 & 62.83 & 92.22 & 54.48 & 90.64 & 50.58 \\

    \hline
    LMD $w/o$ RRL & 93.17 & 56.73 & 92.20 & 53.93 & 90.64 & 49.78 \\
    LMD $w/o$ ICC & 94.35 & 61.24 & 93.58 & 52.06 & 91.29 & 46.41 \\
    LMD $w/o$ VFC & 94.62 & 54.46 & \underline{93.75} & 50.24 & 92.06 & 44.82 \\
    LMD $w/o$ FDC & 94.56 & 55.07 & 93.70 & 51.92 & 91.73 & 46.23 \\

    \textbf{LMD} & \textbf{95.11} & \textbf{70.75} & \textbf{94.01} & \textbf{59.39} & \textbf{93.69} & \textbf{56.88} \\
\toprule
\end{tabular}
}
\end{table*}

\subsection{Comparisons on the Hyper-Kvasir Dataset}
Following \citep{li2022flat, ju2022flexible, fang2023revisiting}, We evaluate our LMD frameworks and other approaches on the \emph{Hyper-Kvasir} dataset using the metrics including Area under the ROC curve (AUC), balanced accuracy (BACC), macro F1 score (F1), quadratic weighted kappa (Kappa), macro Precision (Precision), and macro Recall (Recall). As shown in Table \ref{tab:kvasir}, our LMD framework is superior to state-of-the-art approaches on all evaluation metrics, with a particularly noteworthy balanced accuracy of 67.27\%, demonstrating its ability for unbiased classification on an imbalanced dataset. 
Compared to the state-of-the-art reweighting approach Seesaw loss \citep{wang2021seesaw}, our LMD framework achieves an increase in AUC of 0.89\%, an 8.14\% increase in BACC, a 3.50\% increase in F1, a 7.44\% increase in quadratic weighted kappa, and a 1.67\% increase in macro precision. 
Our LMD framework also outperforms the state-of-the-art two-stage method in computer vision, CC-SAM \citep{zhou2023class}, with a 0.12\% increase in AUC, a 3.87\% increase in BACC, a 1.27\% increase in F1, a 5.24\% increase in Kappa, and a 0.75\% increase in Precision.
Compared to FCD \citep{li2022flat}, the most recent study on long-tailed medical datasets, our LMD framework achieves a 2.84\% increase in AUC, a 5.36\% increase in BACC, a 5.71\% increase in F1, a 7.30\% increase in Kappa, and a 4.90\% increase in Precision.
Notably, our LMD framework exceeds the decoupling method \citep{kang2019decoupling} by a large margin: 4.30\% in AUC, 6.91\% in BACC, 2.46\% in F1, 6.36\% in Kappa, and 1.43\% in Precision, demonstrating the effectiveness of the proposed modules.

\begin{figure}[t!]
\centering
\subfigure[Class Distribution]{
\includegraphics[width=.49\textwidth]{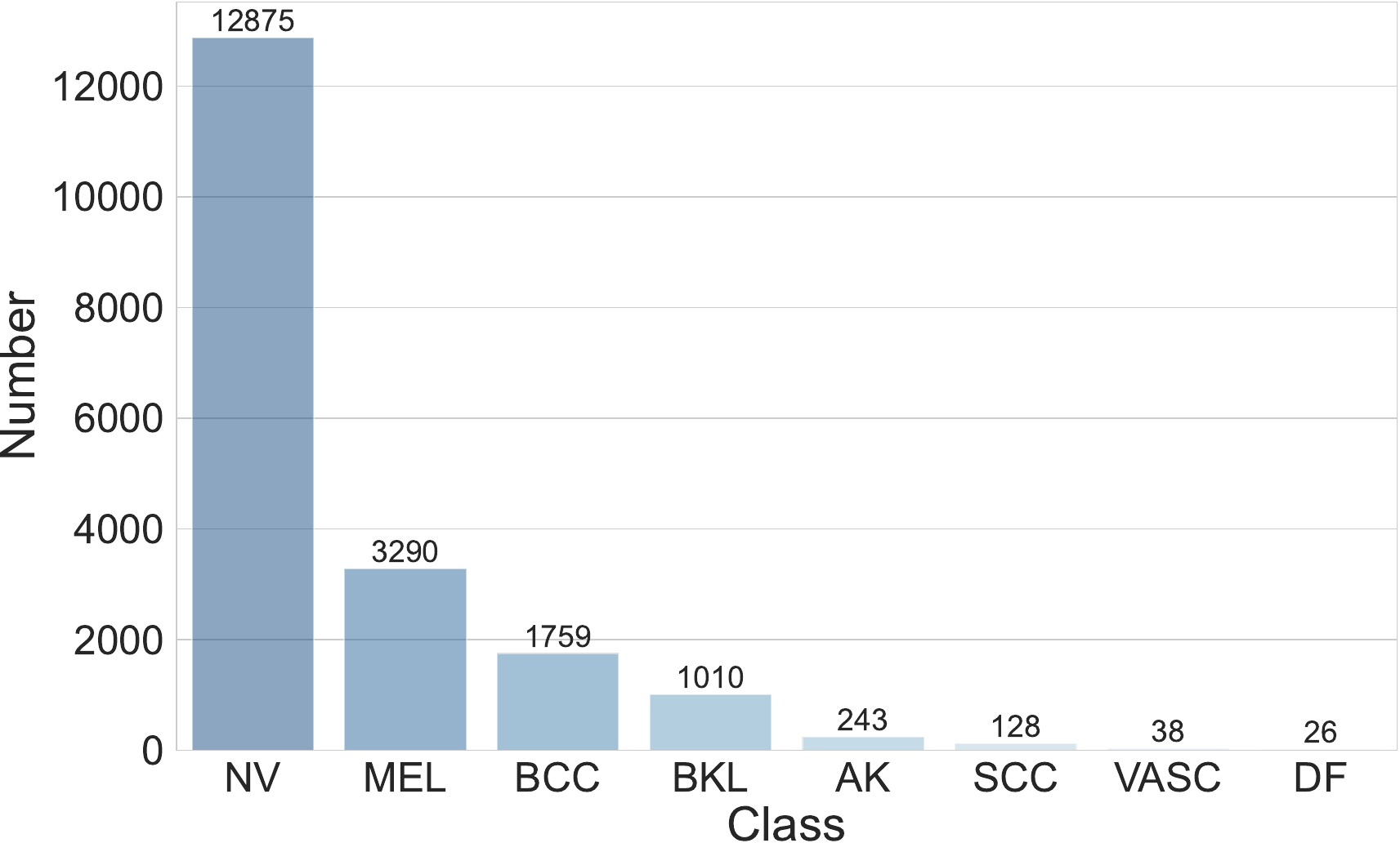}}
\subfigure[Class-wise Recall Rates]{
\includegraphics[width=.49\textwidth]{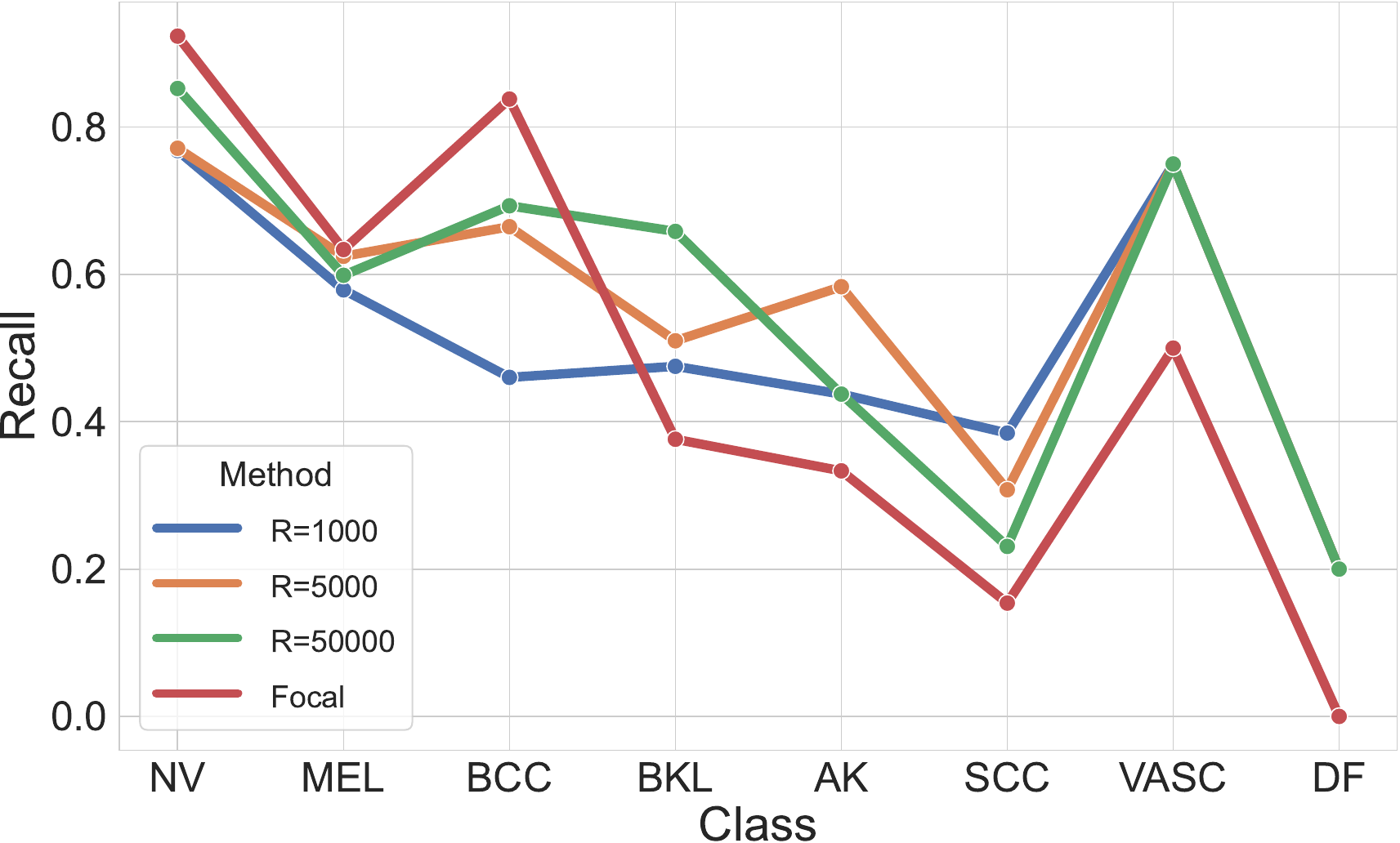}}
\caption{Ablation study of the resampling size $R$ of the VFC on the \emph{ISIC-2019-LT} dataset at $r=500$.}
\vspace{-5pt}
\label{fig:class_isic19}
\end{figure}

\subsection{Comparisons on the ISIC-Archive-LT Dataset}
We compare our LMD with leading approaches using a more challenging dataset, specifically the \emph{ISIC-Archive-LT}. We utilize Balanced Accuracy (BACC) as a measure to assess the classification performance across various class groups, which include the head, medium, and tail classes, as well as the overall BACC across all classes.
As illustrated in Table \ref{tab:archive}, our LMD framework outperforms others by achieving the highest balanced accuracy across medium, tail, and overall classes, indicating its superior performance in balanced classification on the long-tailed dataset. 
When compared with the second-best method, CC-SAM \citep{zhou2023class}, our LMD framework demonstrates a significant improvement, with a 6.02\% increase in the BACC of medium classes, a 7.99\% increase in the BACC of tail classes, and a 4.02\% increase in the BACC of overall classes.
our LMD framework also surpasses the performance of the state-of-the-art two-stage method, CICL \citep{marrakchi2021fighting}, by achieving a 9.11\% increase in the BACC of medium classes, an impressive 25.58\% increase in the BACC of tail classes, and a 9.98\% increase in the BACC of overall classes.
In comparison to the standard decoupling method \citep{kang2019decoupling}, our LMD framework exhibits a significant improvement, with a 12.66\% increase in the BACC of medium classes, a remarkable 29.41\% increase in the BACC of tail classes, and a 13.73\% increase in the BACC of overall classes.
Notably, our LMD framework also achieves the most balanced classification across head, medium, and tail classes. The difference between the BACC of head and medium classes is only 2.74\%, and between the BACC of head and tail classes, it is just 3.20\%. This demonstrates the capability of our LMD framework to balance the classification across all categories.

\begin{figure*}[t!]
\centering
\subfigure[Decoupling \citep{kang2019decoupling}]{
\label{fig:tsne1}
\includegraphics[width=.32\textwidth]{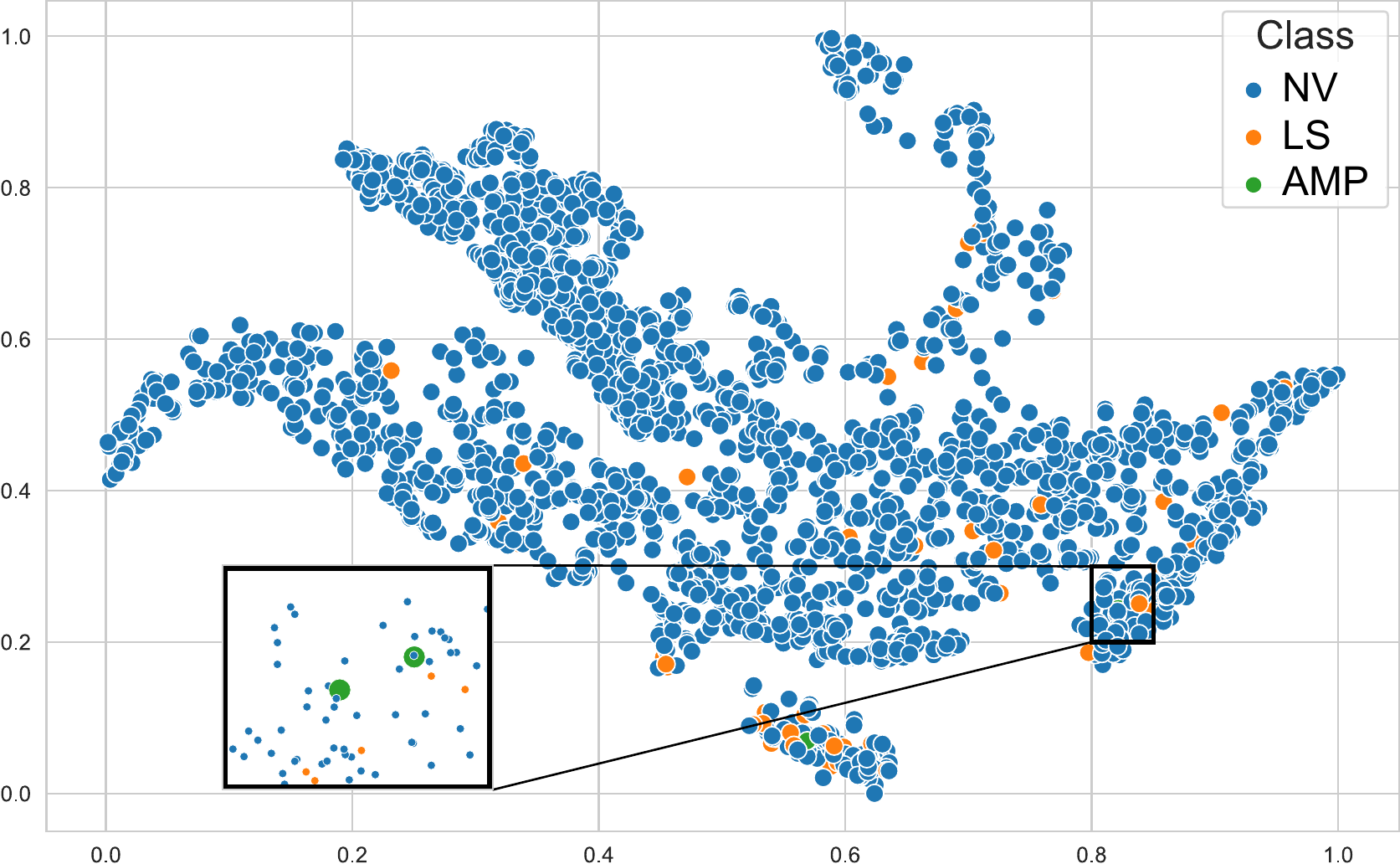}}
\subfigure[LMD $w/o$ VFC]{
\label{fig:tsne2}
\includegraphics[width=.32\textwidth]{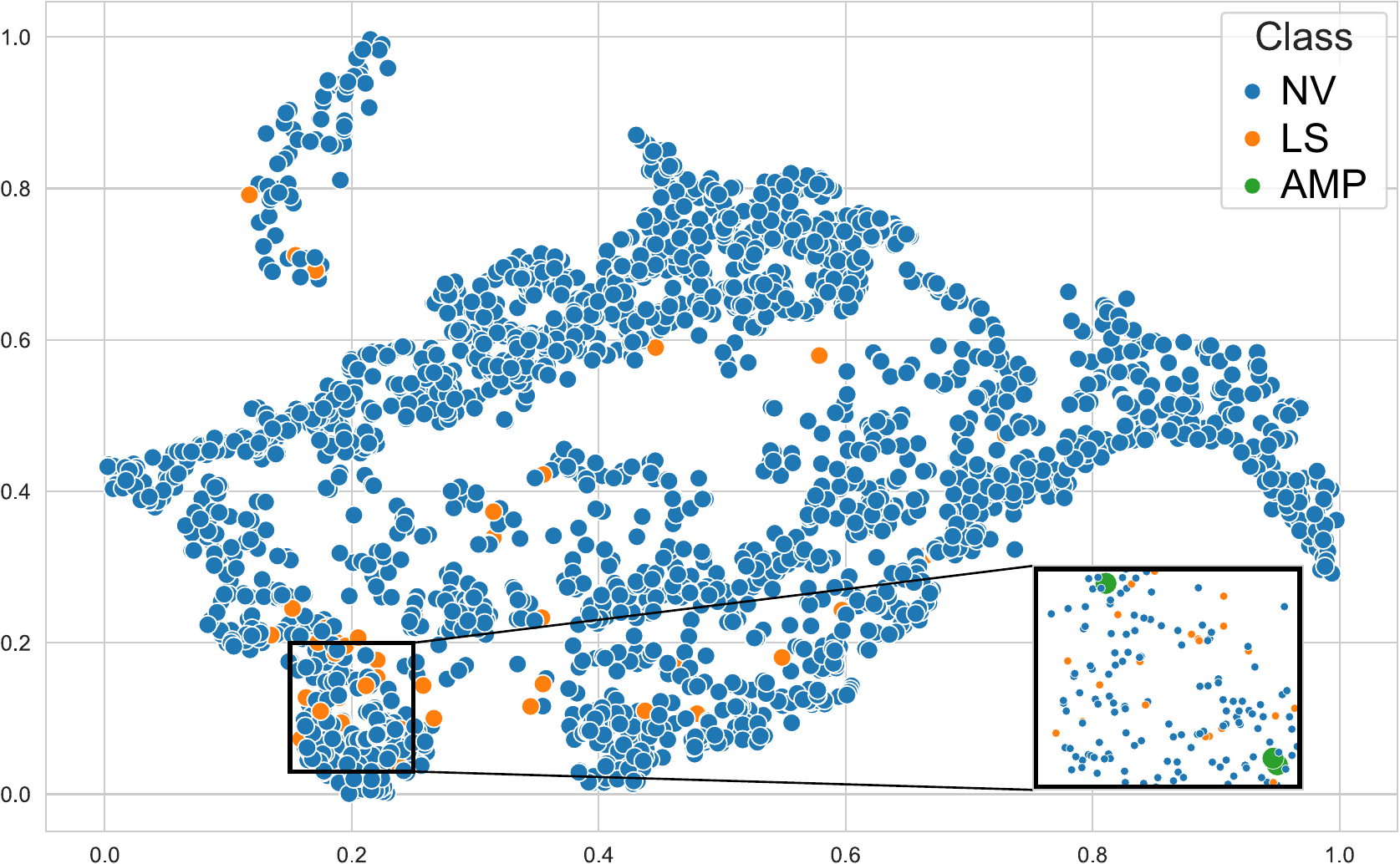}}
\subfigure[LMD]{
\label{fig:tsne3}
\includegraphics[width=.32\textwidth]{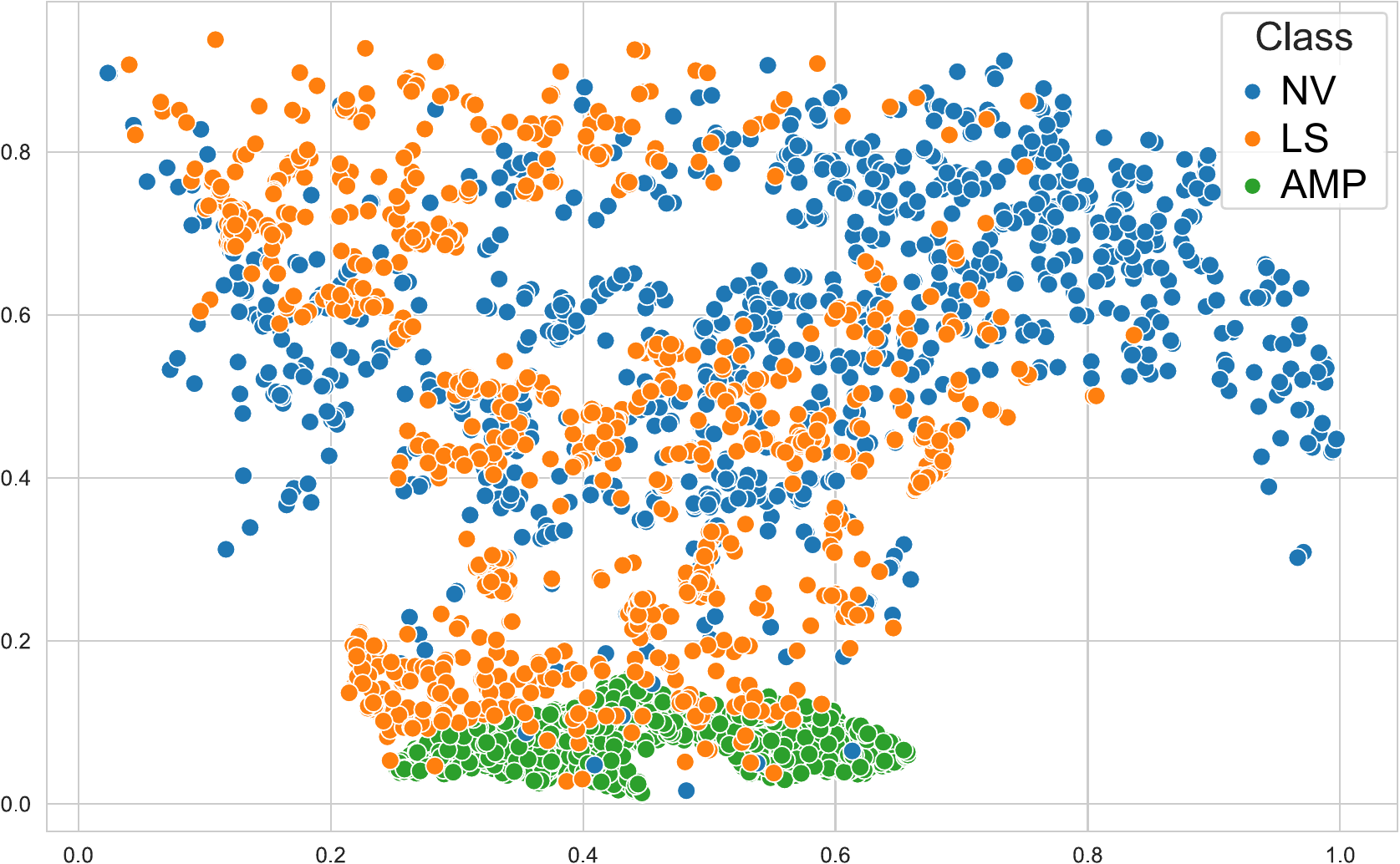}}
\caption{Visualization of feature representations using (a) Decoupling, (b) our LMD framework without the VFC module, and (c) our LMD framework on the head, medium, and tail classes of the \emph{ISIC-Archive-LT} dataset. Our LMD framework demonstrates a clearer and more balanced clustering.}
\label{fig:tsne_hmt}
\end{figure*}

\begin{figure*}[t!]
\centering
\subfigure[Decoupling \citep{kang2019decoupling}]{
\label{fig:tsne4}
\includegraphics[width=.32\textwidth]{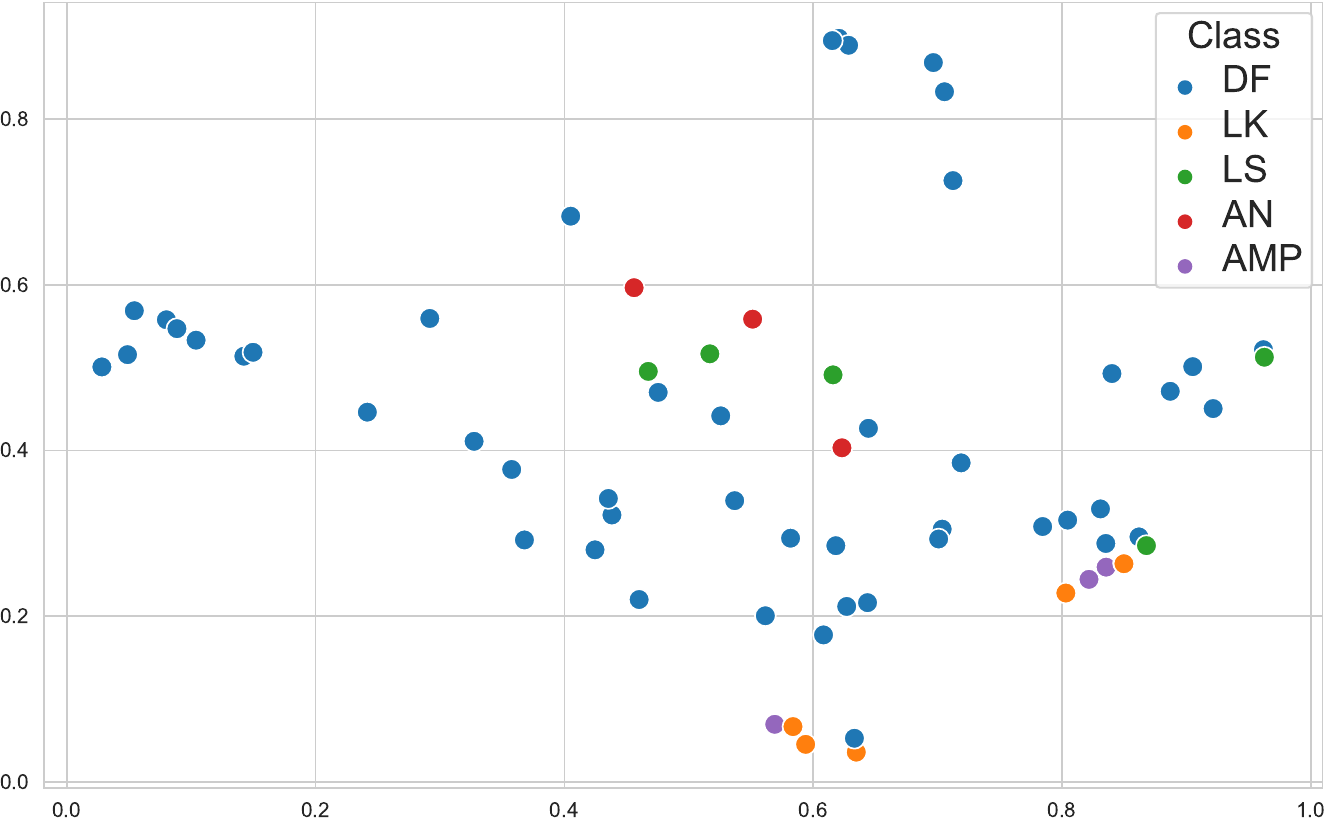}}
\subfigure[LMD $w/o$ VFC]{
\label{fig:tsne5}
\includegraphics[width=.32\textwidth]{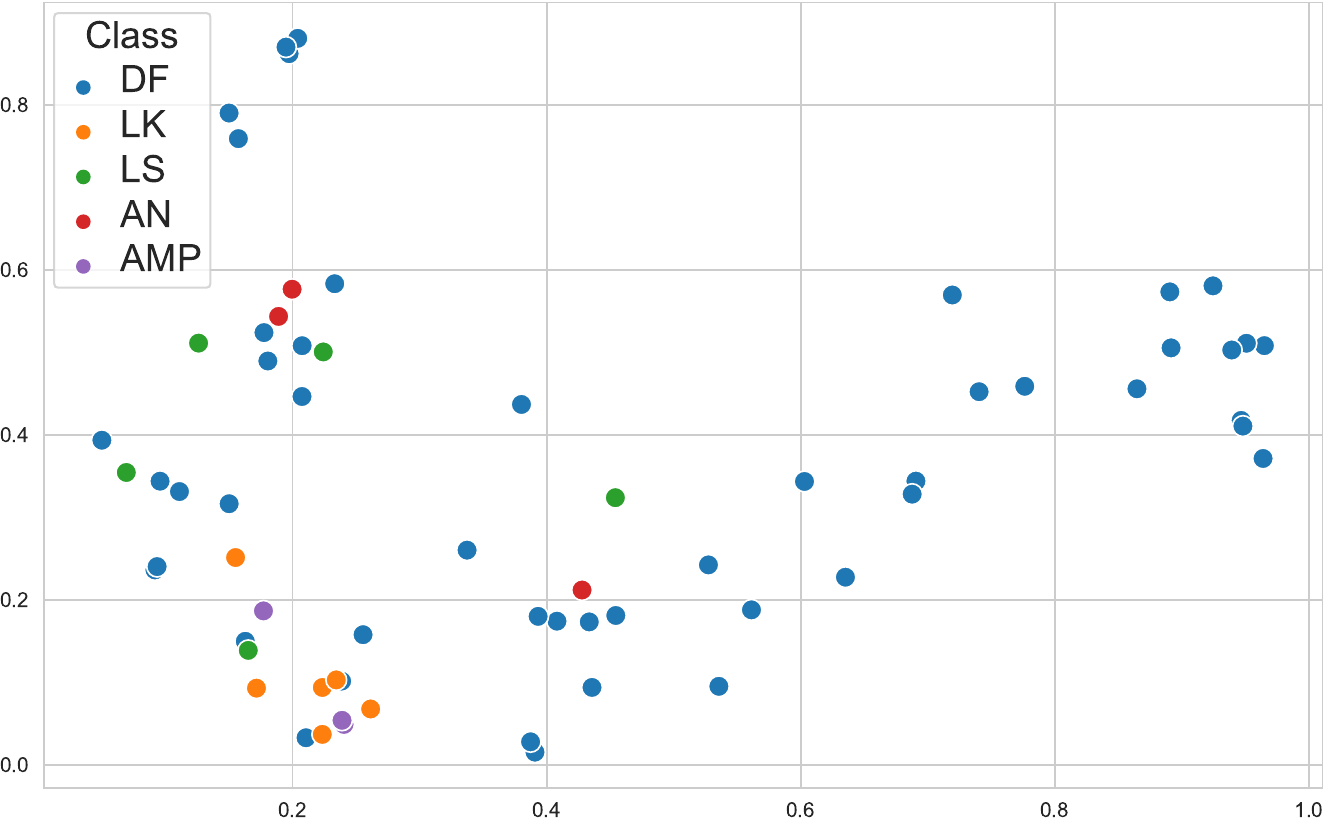}}
\subfigure[LMD]{
\label{fig:tsne6}
\includegraphics[width=.32\textwidth]{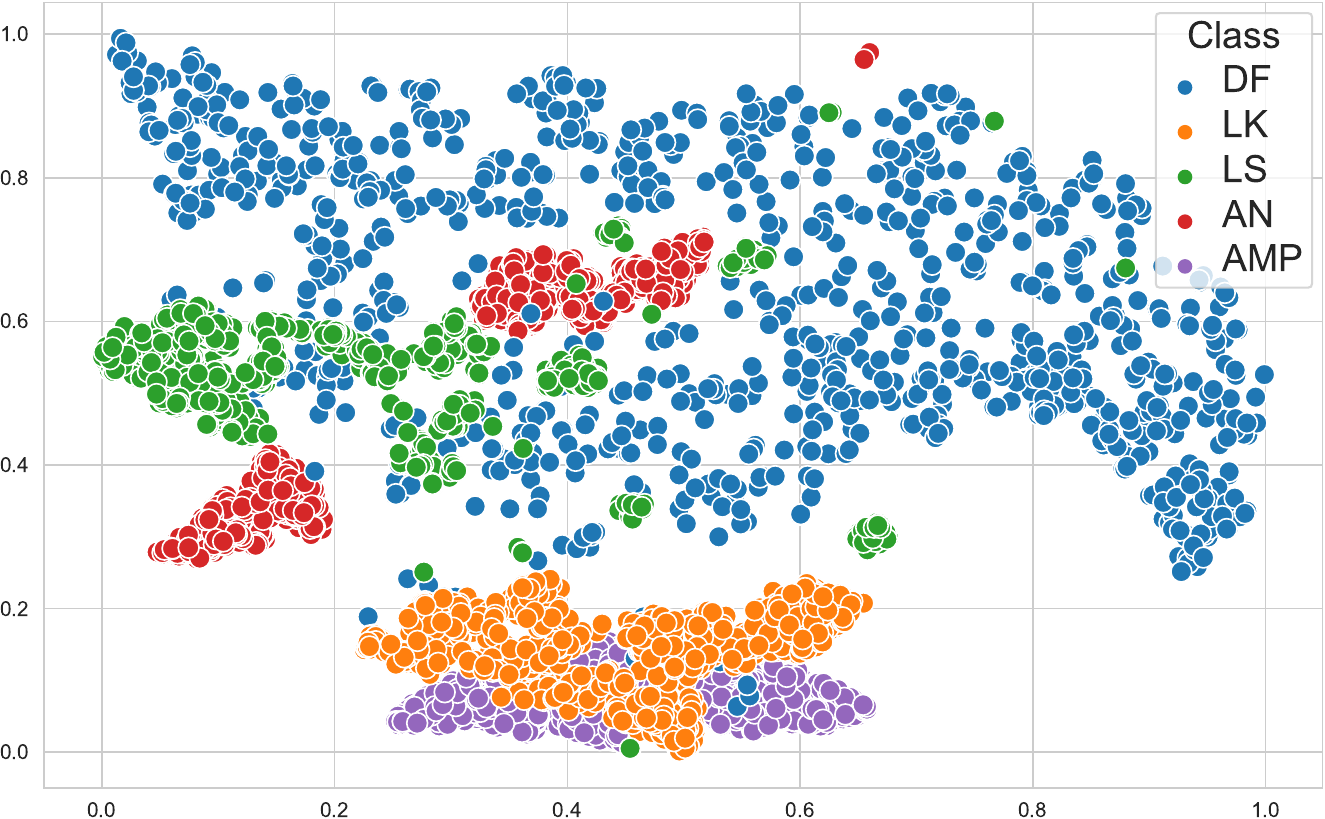}}
\caption{Visualization of feature representations using (a) Decoupling, (b) our LMD framework without the VFC module, and (c) our LMD framework on the tail classes of the \emph{ISIC-Archive-LT} dataset.}
\label{fig:tsne_tail}
\end{figure*}

\subsection{Comparisons on the ISIC-2019-LT Dataset}
We evaluate all the approaches on the \emph{ISIC-2019-LT} dataset under different imbalance factors. As shown in Table \ref{tab:isic19}, our LMD significantly outperforms other methods, achieving an AUC of 95.11\%, 94.01\%, and 93.69\%, as well as a BACC of 70.75\%, 59.39\%, and 56.88\% under imbalance factors $r$ of 100, 300, and 500, respectively.
Compared to the leading long-tailed study, CC-SAM \citep{zhou2023class}, our LMD framework realizes an increase of 0.91\% in AUC and 5.39\% in BACC at $r=100$, 3.30\% in AUC and 4.13\% in BACC at $r=300$, and 4.24\% in AUC and 4.22\% in BACC at $r=500$.
our LMD framework also outperforms the decoupling method \citep{kang2019decoupling} by a 0.14\% increase in AUC and a 5.19\% increase in BACC at $r=100$, a 1.22\% increase in AUC and a 5.50\% increase in BACC at $r=300$, and a 2.78\% increase in AUC and a 4.44\% increase in BACC at $r=500$.
Compared to the cutting-edge resampling approach of the medical long-tailed study, Bal-Mixup \citep{galdran2021balanced}, our LMD framework achieves an increase of 1.27\% in AUC and an increase of 8.84\% in BACC at $r=100$, a 0.61\% improvement in AUC and a 9.73\% improvement in BACC at $r=300$, and an increase of 3.08\% in AUC and a remarkable 16.11\% in BACC at $r=500$, illustrating the effectiveness of our LMD framework.

\begin{figure}[t!]
\centering
\subfigure[\emph{ISIC-Archive-LT}]{
\label{fig:em_archive}
\includegraphics[width=.32\textwidth]{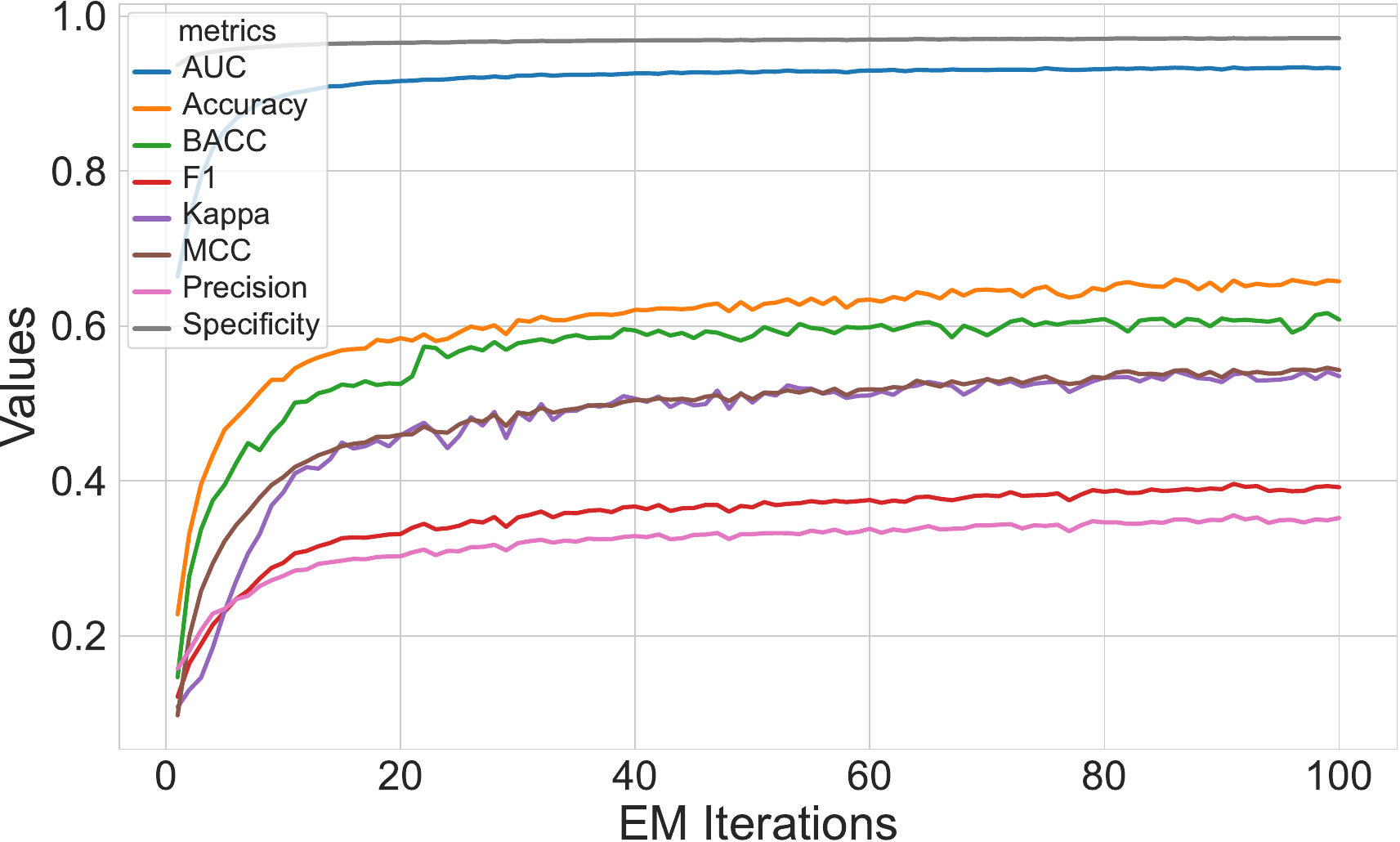}}
\subfigure[\emph{Hyper-Kvasir}]{
\label{fig:em_kvasir}
\includegraphics[width=.32\textwidth]{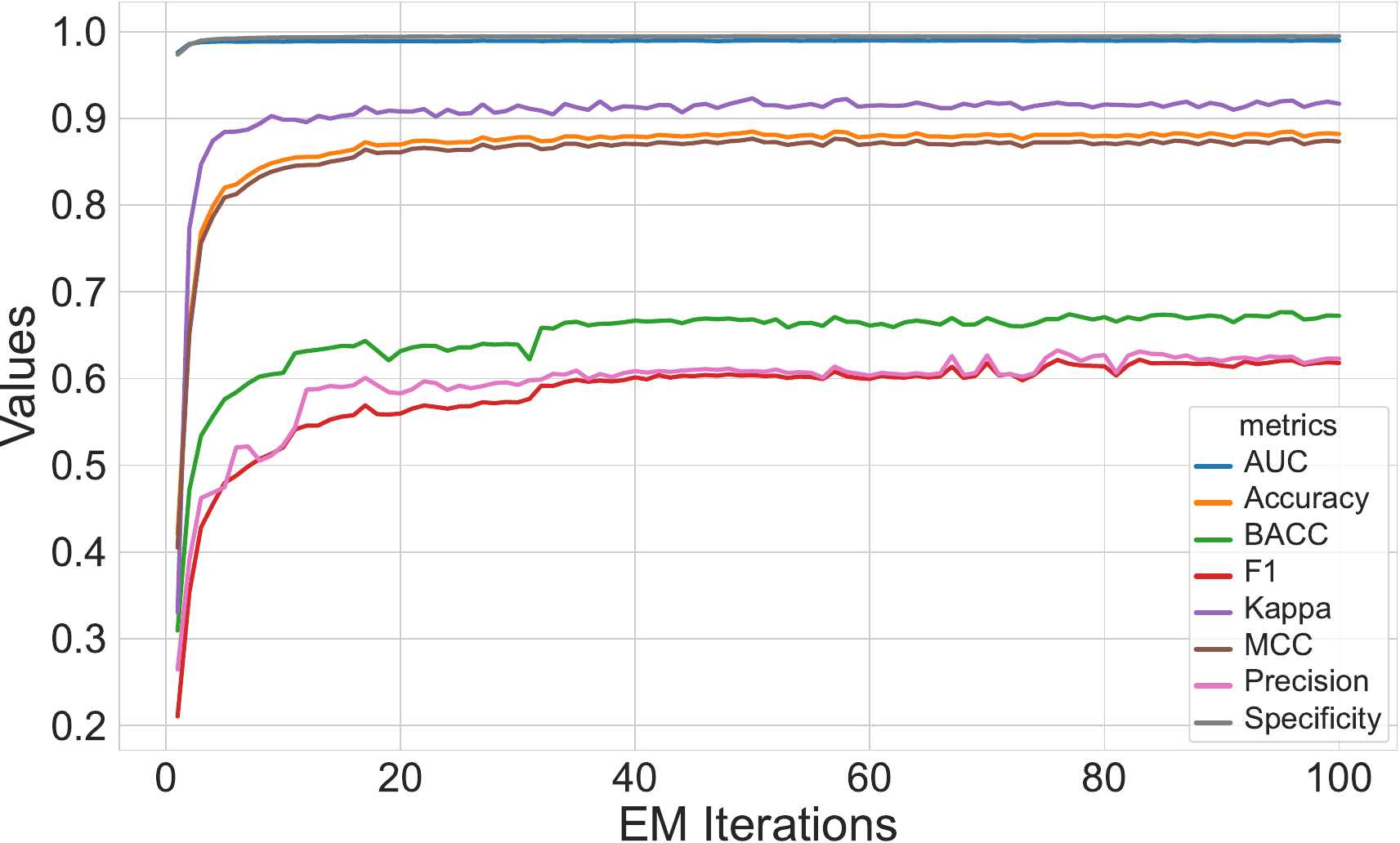}}
\subfigure[\emph{Computational overhead}]{
\label{fig:em_gpu}
\includegraphics[width=.32\textwidth]{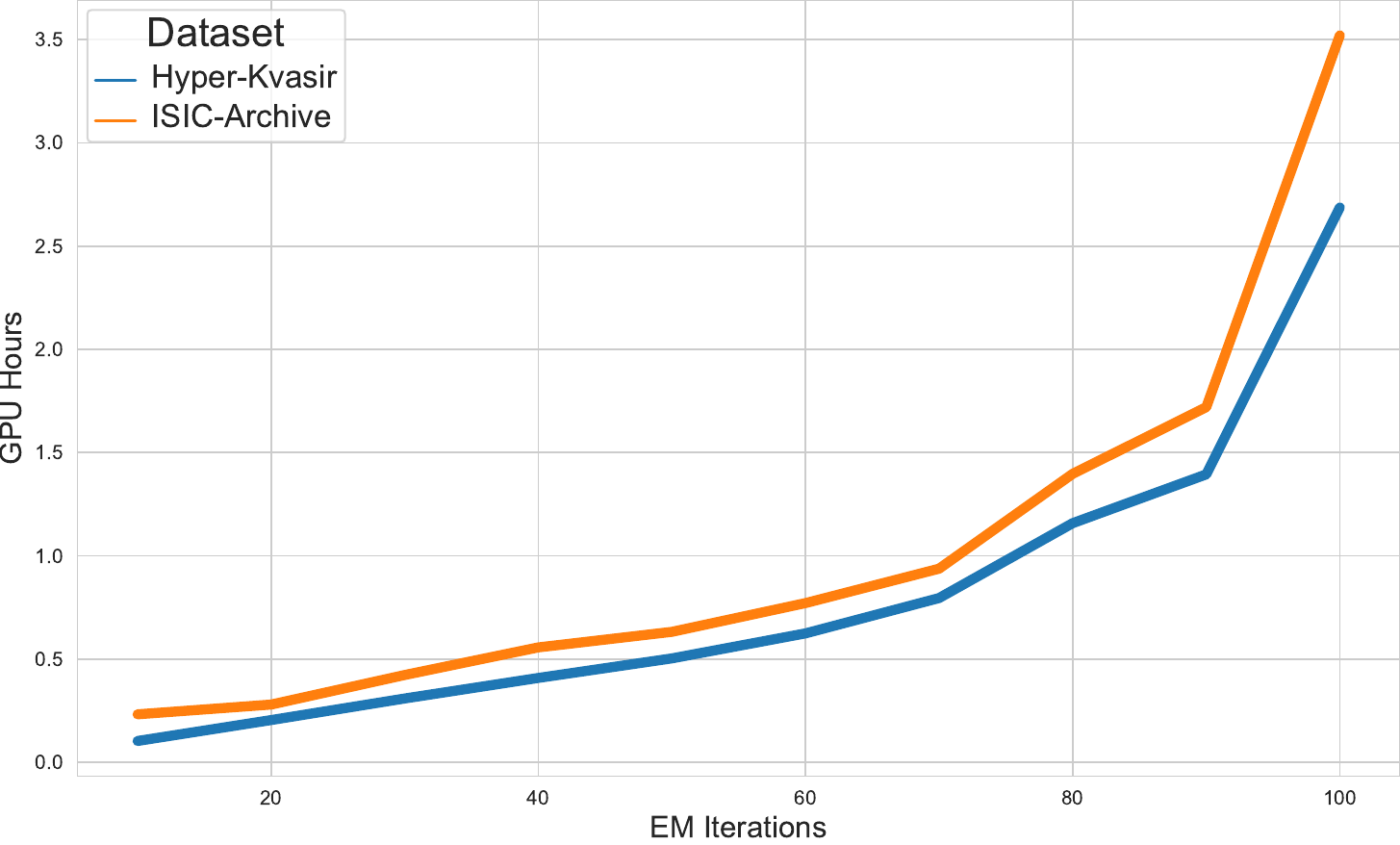}}
\caption{Ablation study of the number of iterations $J$ at ICC.}
\vspace{-5pt}
\label{fig:em}
\end{figure}

\subsection{Ablation Study}
As shown in Table \ref{tab:kvasir}, \ref{tab:archive} and \ref{tab:isic19}, to verify the effectiveness of the RRL, ICC, and VFC modules, we conduct an ablation study on all presented long-tailed datasets. \pl{Specifically, we individually disable the RRL (referred to as LMD $w/o$ RRL), the ICC (referred to as LMD $w/o$ ICC), the VFC (referred to as LMD $w/o$ VFC), and the FDC (referred to as LMD $w/o$ FDC) as the baselines.} In more detail, disabling the MRC results in a 2.97\% decrease in AUC, a 6.08\% decrease in BACC, a 3.36\% decrease in F1, a 2.50\% decrease in Kappa and a 3.80\% decrease in Precision, as shown in Table \ref{tab:kvasir}, which indicates the effectiveness of the MRC module in improving the representation ability of the encoder.
As shown in Table \ref{tab:archive}, disabling the VFC results in an 8.53\% increase in the BACC of head classes, an 11.81\% decrease in the BACC of medium classes, a 22.14\% decrease in the BACC of tail classes, and an 8.07\% decrease in the BACC of overall classes, illustrating the effectiveness of the VFC module in balancing the classification. As demonstrated in Table \ref{tab:isic19}, disabling the ICC results in a 0.76\% decrease in AUC and a 9.51\% decrease in BACC at $r=100$, a 0.43\% decrease in AUC and a 7.33\% decrease in BACC at $r=300$, and a 2.40\% decrease in AUC and a 10.47\% decrease in BACC at $r=500$. \pl{Finally, as indicated in Table \ref{tab:kvasir}, disabling the FDC module leads to decreases in AUC, F1, Kappa, Precision, and Recall by 0.28\%, 1.48\%, 4.45\%, 1.93\%, and 3.38\%, respectively, further demonstrating the effectiveness of the proposed FDC module.}

\begin{table}[t!]
\caption{Ablation study of loss weight $\lambda_e$ on the \emph{Hyper-Kvasir} dataset.\label{tab:lmd}}
\centering
\begin{adjustbox}{center}
\begin{threeparttable}
%\scalebox{0.86}{
\begin{tabular}{@{\extracolsep{\fill}}c| ccccc}
\toprule
\bottomrule
$\lambda_e$ & AUC (\%) & BACC (\%) & F1 (\%) & Kappa (\%) & Precision (\%) \\
\hline
$10^{-2}$ & 97.88 & 55.90 & 50.99 & 86.27 & 54.47 \\
$10^{-3}$ & 98.33 & 60.60 & 57.47 & 86.32 & 56.82 \\
$10^{-4}$ & \textbf{98.96} & \textbf{67.27} & \textbf{62.99} & \textbf{92.43} & \textbf{62.35} \\
$10^{-5}$ & 98.42 & 62.82 & 60.60 & 88.87 & 59.89 \\
$10^{-6}$ & 98.68 & 60.39 & 58.29 & 87.57 & 57.84 \\

\toprule
\end{tabular} 
% }
\vspace{-8pt}
\end{threeparttable}
\end{adjustbox}
\end{table}

\begin{figure}[t!]
\centering
\includegraphics[width=.46\textwidth]{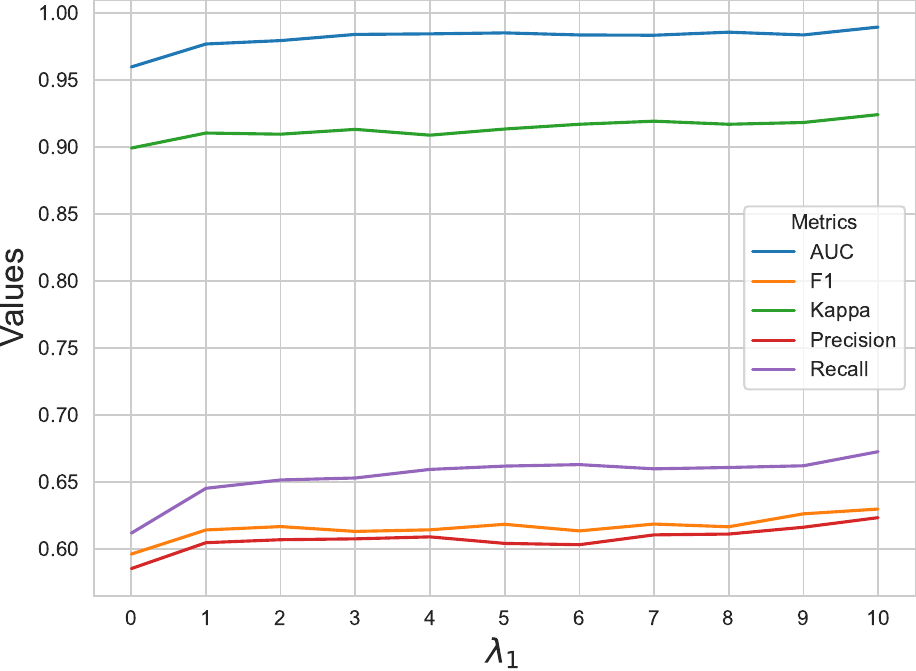}
\caption{\pl{Ablation study of the parameter $\lambda_1$ on the \emph{Hyper-Kvasir} dataset.}}
\vspace{-5pt}
\label{fig:lambda1}
\end{figure}

Moreover, to demonstrate the impact of the VFC module in a more intuitive manner, we visualize the feature representations using t-SNE \citep{van2008visualizing} for different classes. We compare the distribution of feature representations obtained with the decoupling method \citep{kang2019decoupling}, our LMD framework with the VFC module disabled, and our complete framework.
In Fig. \ref{fig:tsne_hmt}, we present the t-SNE visualization of feature representations for three classes: Nevus (NV), Lentigo Simplex (LS), and Atypical Melanocytic Proliferation (AMP), which are selected from the head, medium, and tail categories of the \emph{ISIC-Archive-LT} dataset, respectively. Fig. \ref{fig:tsne3} showcases that our LMD framework with Virtual Feature Compensation is able to achieve a more balanced distribution and a clearer clustering across the head, medium, and tail classes, compared to the decoupling method. 
In Fig. \ref{fig:tsne_tail}, we further visualize the feature representation distribution of the tail classes, which include the Dermatofibroma (DF), Lichenoid Keratosis (LK), Lentigo Simplex (LS), Angioma (AN), and Atypical Melanocytic Proliferation (AMP). Our LMD framework with the VFC module can produce abundant virtual features for each class with an equal number, facilitating a more balanced and distinct clustering for tail classes, as depicted in Fig. \ref{fig:tsne6}. Notably, even with a limited sample size, our LMD framework without the VFC still achieves a more distinct clustering compared to the standard decoupling method, as demonstrated in Fig. \ref{fig:tsne2} and \ref{fig:tsne5}, which demonstrates the efficacy of the MRC module.

\subsection{Hyper-Parameter Analysis}
We evaluate the impact of different hyper-parameters on various datasets. 
\pl{In Table \ref{tab:lmd}, we evaluate the impact of the weight $\lambda_e$ of the distribution loss at the Expectation step. The results suggest that $\lambda_e = 10^{-4}$ is the optimal choice compared to other values. Consequently, we have set $\lambda_e = 10^{-4}$ for subsequent experiments.}
\pl{As shown in Table \ref{tab:bal}, class-balanced sampling during the Maximization step positively influences classification balance, confirming that impartial estimation is essential for achieving a more accurate multivariate Gaussian distribution of features, which subsequentially benefits the VFC and classifier calibration. Conversely, using balanced sampling during the Expectation step leads to negative outcomes, indicating inadequate representation learning on the encoder due to a reduced number of head class samples.}
\pl{As depicted in Fig. \ref{fig:class_isic19}, our LMD framework with different resampling sizes $R$ achieves higher BACC on the medium and tail classes, compared to the baseline method. However, as shown, the optimal selection of $R$ is often empirical. Different values of $R$ yield varying classification performances for specific classes, highlighting the need for a case-specific selection strategy. In this study, we set $R$ as 50,000 for all experiments.} 
We also evaluate the impact of the number of iterations at the ICC. As shown in Fig. \ref{fig:em}, increasing the number of interactions resulted in improved metrics, demonstrating the ICC's performance in approaching the global optimum. \pl{Furthermore, Fig. \ref{fig:em_gpu} reveals that the computational overhead (measured in GPU hours) increases with the number of iterations, highlighting the necessity of strategically selecting the value of $J$ to balance between performance and computational cost. Notably, this additional overhead is incurred solely during the training phase, as the model directly processes the input during inference without the necessity of generating virtual features.}
\pl{As shown in Fig. \ref{fig:lambda1}, increasing the loss weight term $\lambda_1$ in the first stage improves performance in the \emph{Hyper-Kvasir} dataset, further showcasing the effectiveness of the proposed RRL module.}
\pl{As shown in Fig. \ref{fig:tsne_hmt}, Fig. \ref{fig:tsne_tail}, and Fig. \ref{fig:em}, compared to existing methods \citep{kang2019decoupling, wang2021seesaw}, our LMD framework achieves a more adequate and balanced feature distribution while demonstrating improved performance with an increasing number of iterations, thus outperforming existing methods.} 

\begin{table}[t!]
\caption{Ablation study of class-balanced sampling on \emph{ISIC-Archive-LT}. $\triangle$ denotes uniform sampling and $\Circle$ denotes class-balanced sampling.\label{tab:bal}}
\centering
\begin{adjustbox}{center}
\begin{threeparttable}
%\scalebox{0.86}{
\begin{tabular}{@{\extracolsep{\fill}}l cc | cccc}
\toprule
\bottomrule
& \multirow{2}{*}{E-Step} & \multirow{2}{*}{M-Step} & \multicolumn{4}{c}{BACC} \\
\cline{4-7}
& & & Head & Medium & Tail & Overall \\
\hline
1 & $\triangle$ & $\triangle$ & \textbf{63.94 $\pm$ 0.13} & 61.00 $\pm$ 0.10 & 56.79 $\pm$ 0.34 & 60.04 $\pm$ 0.22 \\
2 & $\Circle$ & $\triangle$ & 63.22 $\pm$ 0.12 & 59.89 $\pm$ 0.09 & 54.77 $\pm$ 0.32 & 58.65 $\pm$ 0.22 \\
3 & $\Circle$ & $\Circle$ & 63.54 $\pm$ 0.11 & 60.39 $\pm$ 0.11 & 54.01 $\pm$ 0.30 & 58.55 $\pm$ 0.20 \\
4 & $\triangle$ & $\Circle$ & 63.79 $\pm$ 0.14 & \textbf{61.05 $\pm$ 0.12} & \textbf{60.59 $\pm$ 0.31} & \textbf{61.63 $\pm$ 0.21} \\
\toprule
\end{tabular} 
%}
%\vspace{-5pt}
\end{threeparttable}
\end{adjustbox}
\end{table}

\section{Conclusion}
To tackle the challenges posed by long-tail issues in computer-aided diagnosis, we devise the LMD framework aimed at enhancing medical image classification through a two-stage process. At first, we devise the Relation-aware Representation Learning technique to boost the encoder's representation capabilities by incorporating multi-view relation-aware consistency. Subsequently, we present the Iterative Classifier Calibration method, which trains an unbiased classifier by generating numerous virtual features and iteratively refining both the encoder and classifier. Comprehensive experiments conducted on three long-tailed medical datasets validate the effectiveness of the LMD framework, which significantly surpasses the performance of current leading algorithms.

\section*{CRediT authorship contribution statement}
\textbf{Li Pan:} Conceptualization, Methodology, Software,
Validation, Writing – original draft. 
\textbf{Yupei Zhang:} Conceptualization, Methodology, Software, Validation, Writing – original draft. 
\textbf{Qiushi Yang:} Formal analysis, Visualization, Investigation, Writing – review \& editing. 
\textbf{Tan Li:} Formal analysis, Resources, Visualization, Investigation, Writing – review \& editing. 
\textbf{Zhen Chen:} Methodology, Investigation, Supervision, Writing – review \& editing.

\section*{Declaration of Competing Interest}
The authors declare that they have no known competing financial interests or personal relationships that could have appeared to influence the work reported in this paper.

% \section*{Acknowledgments}
% This work was supported in part by the InnoHK program.





\bibliographystyle{elsarticle-harv.bst}
\bibliography{refs}

\end{document}